%% file: egpaper_final.tex
\documentclass[10pt,twocolumn,letterpaper]{article}

\usepackage[accsupp]{axessibility}  

\usepackage{iccv}
\usepackage{times}
\usepackage{epsfig}
\usepackage{graphicx}
\usepackage{amsmath}
\usepackage{amssymb}
\usepackage{tabularx}

\usepackage{booktabs}
\usepackage{duckuments}
\usepackage{enumitem}
\usepackage{multirow}
\usepackage{subcaption}
\usepackage{pifont}
\usepackage[normalem]{ulem}

\usepackage{xspace}
\usepackage[dvipsnames]{xcolor}

\newcommand{\Fref}[1]{Figure \ref{#1}}
\newcommand{\fref}[1]{Figure \ref{#1}}
\newcommand{\Sref}[1]{$\S$\ref{#1}}

\newcommand{\Tref}[1]{Table \ref{#1}}

\newcommand{\Aref}[1]{Appendix \ref{#1}}

\newcommand{\xmark}{\ding{53}}
\newcommand{\degree}{\ensuremath{^\circ}}

\newcommand{\myparagraph}[1]{\smallskip\noindent{\textbf{#1} {} {}}}

\newcommand{\rad}{R_\bg}  

\newcommand{\fg}{\text{fg}}
\newcommand{\bg}{\text{bg}}  
\newcommand{\fgmlp}{\textit{F}_{\fg}}  
\newcommand{\bgmlp}{\textit{F}_{\bg}}  
\newcommand{\vf}{\mathbf{\phi}}  

\newcommand{\vr}{\mathbf{r}}
\newcommand{\vx}{\mathbf{x}}
\newcommand{\xbg}{\mathbf{x}^{\bg}}  
\newcommand{\sbg}{\mathbf{s}}  

\newcommand{\Nfg}{\mathbf{N}_{\fg}}  

\newcommand{\vw}{\mathbf{w}}  
\newcommand{\vg}{\mathbf{g}}  
\newcommand{\zfg}{\mathbf{z}_{\fg}}  
\newcommand{\zbg}{\mathbf{z}_{\bg}}  
\newcommand{\wbg}{\mathbf{w}_{\bg}}  
\newcommand{\rfg}{\mathbf{\Phi}^{\fg}}  
\newcommand{\rbg}{\mathbf{\Phi}^{\bg}}  

\newcommand{\Ladv}{\mathcal{L}_{\text{adv}}}
\newcommand{\Lr}{\mathcal{L}_{\text{R}_{1}}}

\newcommand{\cmark}{\ding{51}}%

\usepackage{capt-of}
\newcommand\blfootnote[1]{
  \begingroup
  \renewcommand\thefootnote{}\footnote{#1}%
  \addtocounter{footnote}{-1}%
  \endgroup
}

\usepackage[breaklinks=true,colorlinks,bookmarks=false]{hyperref}

\iccvfinalcopy 

\def\iccvPaperID{****} 

\ificcvfinal\fi

\begin{document}

\makeatletter
\newcommand{\settitle}{\@maketitle}
\makeatother

\title{BallGAN: 3D-aware Image Synthesis with a Spherical Background}

\author{Minjung Shin\textsuperscript{1*} \thinspace\quad Yunji Seo\textsuperscript{1} \thinspace\quad Jeongmin Bae\textsuperscript{1} \thinspace\quad Young Sun Choi\textsuperscript{1} \\ Hyunsu Kim\textsuperscript{2} \thinspace\quad Hyeran Byun\textsuperscript{1} \thinspace\quad Youngjung Uh\textsuperscript{1\dag} \\
Yonsei University\textsuperscript{1} \thinspace\quad NAVER AI Lab\textsuperscript{2}\\}

\twocolumn[{
\renewcommand\twocolumn[1][]{#1}
\settitle
\begin{center}
    \centering
    \vspace{-7mm}
    \includegraphics[width=\linewidth]{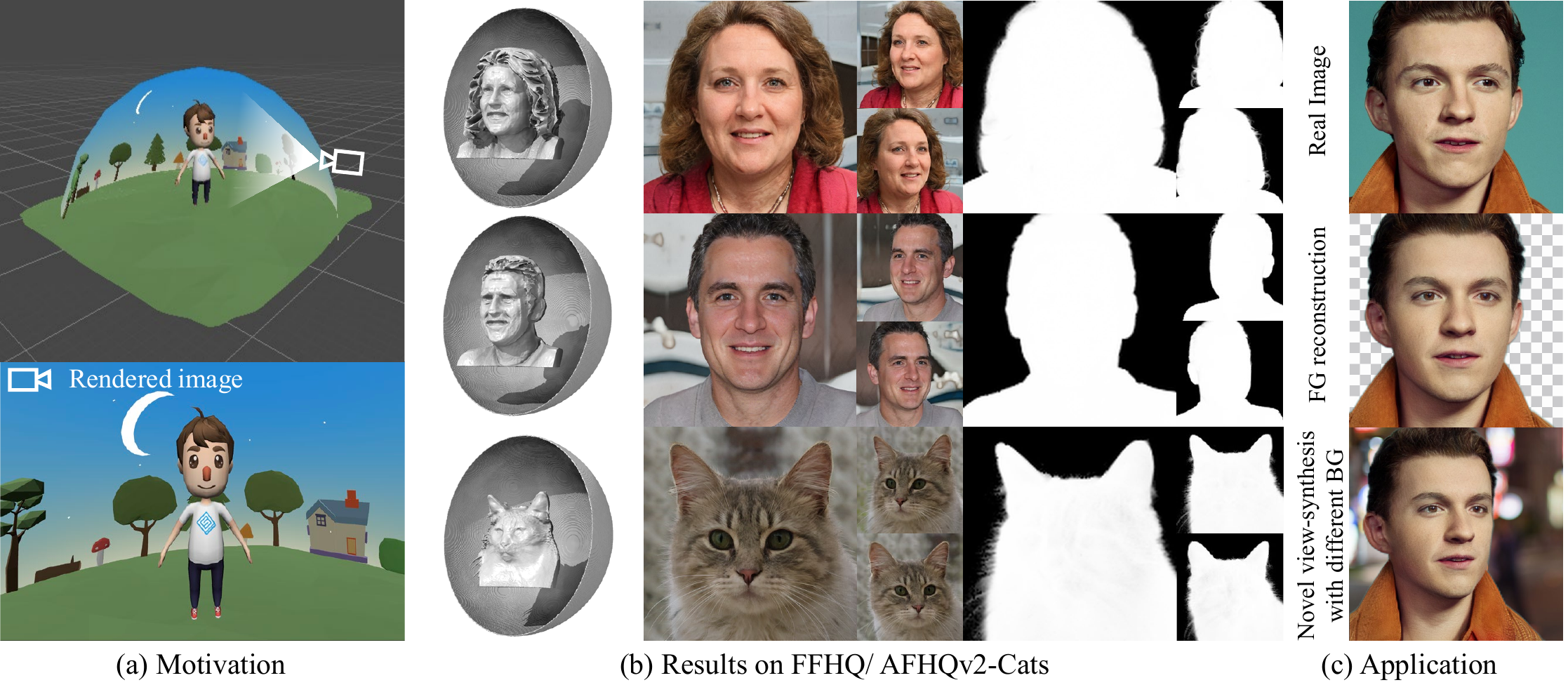}
    \vspace{-8mm}
    \captionof{figure}{(a) 3D space (top) and an image rendered from the white camera (bottom). We are inspired by a 3D graphics technique in which the foreground is represented as a 3D model and the background is approximated as a 2D surface, yet resulting in a realistic appearance on the rendered image. 
    (b) Our method produces high-quality 3D shapes, images, and foreground alpha masks without extra supervision. 
    (c) Realistic novel view rendering on arbitrary backgrounds, even on real image inversion.} 
    \label{fig:teaser}
    \vspace{-3mm}
\end{center}
}]


\begin{abstract}
3D-aware GANs aim to synthesize realistic 3D scenes that can be rendered in arbitrary camera viewpoints, generating high-quality images with well-defined geometry. 
As 3D content creation becomes more popular, the ability to generate foreground objects separately from the background has become a crucial property.
Existing methods have been developed regarding overall image quality, but they can not generate foreground objects only and often show degraded 3D geometry. 
In this work, we propose to represent the background as a spherical surface for multiple reasons inspired by computer graphics. Our method naturally provides foreground-only 3D synthesis facilitating easier 3D content creation. Furthermore, it improves the foreground geometry of 3D-aware GANs and the training stability on datasets with complex backgrounds. Project page: \href{https://minjung-s.github.io/ballgan}{https://minjung-s.github.io/ballgan/}
\vspace{-6mm}
\blfootnote{*Part of the work was done during an internship at NAVER AI Lab.}
\blfootnote{\dag Corresponding author}
\end{abstract}

\section{Introduction}
\label{sec:intro}

Traditional generative adversarial networks (GANs) synthesize realistic images.
Although they provide some control over the camera poses \cite{shen2020interpreting,shen2021closed,harkonen2020ganspace,shoshan2021gan}, they lack explicit 3D understanding of the scenes.
Recently, 3D-aware GANs \cite{nguyen2019hologan,chan2021pi,schwarz2020graf,zhou2021cips} reformulate the generative procedure as modeling the potential 3D scenes and rendering them to images.
The state-of-the-art 3D-aware GANs~\cite{chan2022efficient, gu2021stylenerf,xue2022giraffe} rely on neural radiance fields or their variants to represent 3D scenes.
Note that they can generate 3D scenes even without 3D supervision or multi-view supervision, rendering realistic images across different viewpoints. Although the quality of images generated by 3D-aware GANs continues to improve, their practical usage has been less explored.

\begin{figure}
\begin{center}
    \centering
    \includegraphics[width=\linewidth]{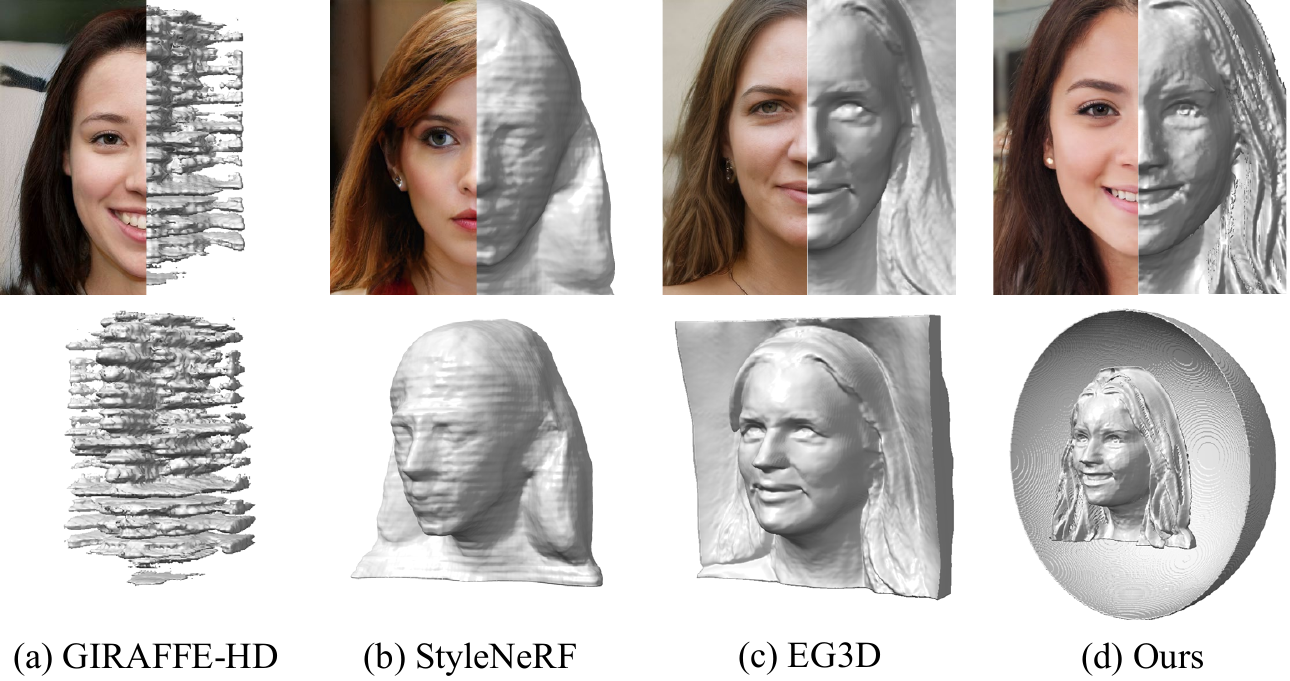}
    \vspace{-5mm}
    \caption{\textbf{Comparison of the 3D geometry extracted by marching cubes.} (a) GIRAFFE-HD exhibits broken 3D shapes, (b) StyleNeRF has jaggy surfaces, and (c) EG3D has hair sticking to the wall. Unlike other models, (d) our model produces high-quality foreground geometry that is separated from the background.} 
    \label{fig:degenerate}
    \vspace{-3mm}
\end{center}
\end{figure}

Solely generating foreground objects is an important element for the practical use of generative models, especially for content creation. 
In this context, the diffusion-based methods have grown popular for 3D object synthesis despite their lack of realism~\cite{jain2022zero,poole2022dreamfusion,lin2022magic3d,singer2023text4d,xu2022dream3d}. 
Some 2D GANs model their output images as a combination of foreground and background, replacing the need for laborious post-processing \cite{bae2022furrygan,bielski2019emergence,zou2022ilsgan}. 
On the other hand, few 3D-aware GANs inadequately separate the background and suffer from broken 3D shapes~\cite{xue2022giraffe} or training instability~\cite{gu2021stylenerf}. 
Objects generated by EG3D~\cite{chan2022efficient} are connected to unrealistic walls as shown in \fref{fig:degenerate}.

Learning to synthesize 3D foreground objects using a single-view dataset is challenging because it lacks both depth and separation supervision.

To solve this problem, we are inspired by a popular approach for video games or movies in the graphics community: representing salient objects with detailed 3D models and approximating peripheral scenery with simple surfaces (\Fref{fig:teaser}a) to reduce the overall complexity. 
Despite approximating the 3D space to 2D, the rendered image achieves a realistic appearance.
We expect the 3D-aware generators with a similar approach to achieve both separation and physically reasonable foreground geometry.

Accordingly, we propose our novel 3D-aware GAN framework, named BallGAN. It approximates the background as a 2D \textit{opaque surface of a sphere} and employs conventional 3D features as the foreground. It accompanies a modified volume rendering equation for the opaque background. In addition, we introduce regularizers for clear foreground geometry and separation.

We demonstrate the strength of our work as follows. By design, BallGAN provides clear foreground-background separation without extra supervision (\Fref{fig:teaser}b). For content creation, it enables inserting generated 3D foregrounds in arbitrary viewpoints without post-processing (\Fref{fig:teaser}c). Our background representation as a spherical surface is generally applicable to any generator architectures or foreground representations. 
BallGAN allows StyleNeRF~\cite{gu2021stylenerf} to be trained on a higher resolution of CompCars\cite{yang2015large}\footnote[1]{StyleNeRF diverges on CompCars while growing from $128^2$ to $256^2$.} and achieve a large FID boost, which is notable as the dataset is challenging due to its complex backgrounds.
More importantly, BallGAN not only enhances multi-view consistency, pose accuracy, and depth reconstruction compared to EG3D, but it also faithfully captures fine details in 3D space that are easy to represent in 2D images but challenging to model in 3D.

\begin{figure*}
\begin{center}
  \centering
  \includegraphics[width=\linewidth]{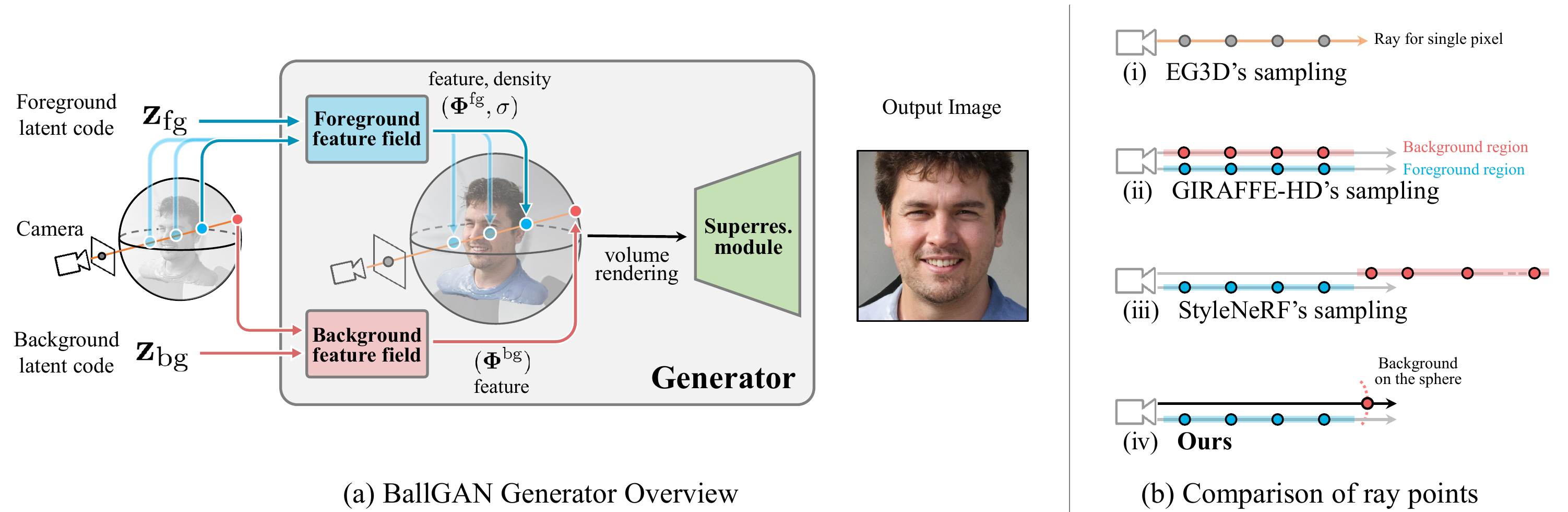}
    \vspace{-5mm}
    \captionof{figure}{\textbf{Overview of the BallGAN generator and definition of ray points.} We bound the 3D space with an opaque background on a spherical surface. (i) EG3D does not separate the background. (ii) GIRAFFE-HD samples the background points within the same range of the foreground. (iii) StyleNeRF samples multiple background points outside the boundary. (iv) We sample a single background point on the sphere. It drastically reduces the depth ambiguity in the background.}
    \label{fig:model}
    \vspace{-3mm}
\end{center}
\end{figure*}

\section{Related work}
\label{sec:relwork}

\myparagraph{Representations for 3D-aware GANs}
Generators in 3D-aware GANs involve representing 3D scenes somehow and rendering them to 2D images so that the generator is aware of the 3D scene given only a collection of unstructured 2D images. HoloGAN~\cite{nguyen2019hologan} represents a scene with a 3D grid of voxels containing feature vectors, \ie, 4D tensor. However, as the 3D grid of voxels is limited by computational complexity, its maximum resolution is $128^2$.

Recent 3D-aware GANs integrate neural radiance fields (NeRFs)~\cite{mildenhall2020nerf}. 
NeRF represents a 3D scene using a coordinate-based function that produces RGB color and density at that coordinates. This 3D scene can be projected onto a 2D image from arbitrary camera poses via volume rendering integral. 
GRAF~\cite{schwarz2020graf} introduces a patch-based discriminator, which dramatically reduces memory usage in high-resolution 3D-aware image synthesis. 
Its successors improve image quality and 3D awareness by 1) enhancing the function for NeRF~\cite{chan2021pi,gu2021stylenerf}, 2) volume rendering feature field followed by neural rendering with upsampling blocks~\cite{gu2021stylenerf,niemeyer2021giraffe,xu20223d,chan2022efficient,xue2022giraffe}, or 3) designing voxel-based~\cite{wu2016learning,gadelha20173d,henzler2019escaping,nguyen2020blockgan,xu20223d}or hybrid~\cite{chan2022efficient} representations. 
Going further, our method introduces a separate NeRF for modeling spherical background, which encloses the foreground of EG3D~\cite{chan2022efficient} or StyleNeRF~\cite{gu2021stylenerf}.

\myparagraph{Scene decomposition}
Some methods decompose the 3D scenes into multiple components. GIRAFFE and its variant \cite{niemeyer2021giraffe,xue2022giraffe} separate scenes into objects and the background, enabling them to control objects independently with the background fixed. However, their background representation lives in the same ray points with the foregrounds, and the 3D geometry does not benefit from the separation. StyleNeRF~\cite{gu2021stylenerf} and EpiGRAF~\cite{skorokhodov2022epigraf} separate the background outside a sphere following NeRF++~\cite{zhang2020nerf++} where the background region goes through the same volume rendering with multiple ray points at variable depth. On the contrary, we remove the depth ambiguity of the background by modeling it with an opaque representation on a 2D spherical surface enclosing the foreground.

\myparagraph{Reducing dimensions} has been a viable option for reducing space and time complexity. TensoRF~\cite{chen2022tensorf} uses a sum of vector-matrix outer products to represent a 3D feature field. EG3D~\cite{chan2022efficient} represents a 3D feature field with three 2D planes to adopt StyleGAN architecture. 
K-Planes~\cite{fridovich2023k} represents a $d$-dimensional scene using $\binom{b}{2}$ planes. While these methods decompose 3D feature fields into low-dimensional feature representations to reduce the memory usage of NeRFs, BallGAN squeezes the background space into a surface to provide an easier task for 3D-aware GANs.

\section{BallGAN}
\label{sec:approach}
In this section, we provide an overview of our framework and describe its key components and intuitions.

\myparagraph{Overview}
We suppose that generating unbounded 3D scenes is too complex to learn relying on a limited guide for producing realistic 2D images. To resolve this challenge, BallGAN bounds the scene in a ball and approximates the background as an opaque spherical surface. We expect it to alleviate the burden of producing correct shapes of the backgrounds because the shape is fixed on a ball.

As shown in \Fref{fig:model}, our generator consists of two backbone networks for foreground and background (\Sref{sec:repr}). Representations from these networks are rendered by our modified volume rendering equation to synthesize images (\Sref{sec:rendering}) and trained with GAN objectives and auxiliary regularizations (\Sref{sec:losses}).

\subsection{Bounding the 3D space}
\label{sec:repr}
While traditional 2D GANs learn to produce arrays of RGB pixels in fixed dimensions, 3D-aware GANs aim to produce realistic images by synthesizing 3D scenes and rendering them into 2D images. 
In contrast to training NeRFs with multi-view observations of a single scene, the only objective for the 3D-aware GANs is producing realistic 2D images. 
In other words, the datasets and the objective functions do not provide any clues for the 3D geometry. To reformulate 3D-aware generation as an easier constrained problem, we approximate the backgrounds on an opaque spherical surface. 

\myparagraph{Background model}
We model the background as a neural feature field defined on a sphere with a fixed radius. 
Given a ray $\vr=\mathbf{o}+t\mathbf{d}$ ($t$ is the distance from the camera center $\mathbf{o}$), we find the 3D background point on the sphere with radius $\rad$ by simply computing the ray's intersection on the sphere surface: 
\begin{equation}
    \xbg=\mathbf{o}+\frac{-2[\mathbf{d}\cdot\mathbf{o}]+\sqrt{(2[\mathbf{d}\cdot\mathbf{o}])^2-4\|\mathbf{d}\|^2(\|\mathbf{o}\|^2-\rad^2)}}{2\|\mathbf{d}\|^2}\mathbf{d}
\end{equation}

Since the background points are on a sphere surface of fixed radius $\rad$, we further reparameterize the 3D coordinates $\vx$ as 2D spherical coordinates $\sbg=(\theta, \phi)$ to further reduce the complexity.

Then we represent the feature field $\bgmlp$ using a StyleGAN2-like architecture :
\begin{equation}
    \bgmlp(\sbg, \zbg) = \vg^n_\vw \circ ... \vg^1_\vw \circ \zeta(\sbg),
\end{equation}
where 
$\vw=\mathbf{f}(\zbg)$ is the style vector produced by a mapping network $\mathbf{f}$ given a noise vector $\zbg$, and $\zeta$ is the positional encoding \cite{tancik2020fourier} of $\sbg$, 
and $\vg_\vw$ denotes $1\times1$ convolutions whose weights are modulated by $\vw$.
Note that there is no mapping for density from the background feature field because our background is an opaque surface. 

Our background representation drastically reduces the number of points to be fed to the model, \ie, only one intersection of our sphere background and the ray $\vr$. Therefore, we do not use hierarchical sampling for the background. 

\Fref{fig:model}b visualizes the difference in space for each method with ray points.
GIRAFFE-HD does not separate the background coordinate space from the foreground, StyleNeRF keeps multiple point candidates for the unbounded continuous depth.
On the other hand, our method separates the foreground and background and bounds the background to lie on the surface. This effectively constrains the solution space and improves training stability and output quality.

\myparagraph{Design choice for background}
One may wonder why we chose the sphere among many alternatives. First, the background should enclose the scene entirely to cover all viewing directions. Thus, an open plane is not available in wide-angle scenes. Second, the background should be identical when observed from all directions to make it easier for the generator to perform consistently well. Therefore, the spherical surface is the only reasonable choice. \Aref{supp:bg design} provides empirical comparison.

\myparagraph{Foreground model}
We adopt StyleNeRF~\cite{gu2021stylenerf} or EG3D~\cite{chan2022efficient} for foreground modeling, where a random foreground code $\zfg$ is fed to StyleGAN2~\cite{Karras2019stylegan2} network to produce implicit or hybrid representation, respectively. 
Formally:
\begin{equation}
(\rfg, \sigma) = \fgmlp(\vx, \zfg).
\end{equation}

Note that our simple and effective background modeling is applicable to arbitrary 3D scene representations other than StyleNeRF and EG3D.

\subsection{Volume rendering}
\label{sec:rendering}
Volume rendering aggregates the neural feature field along the rays through individual pixels to produce feature maps for a given camera pose. 
The conventional volume rendering computes the contribution of all points $\{\vx_i\}$ sampled on a ray using the same equation $T(\vx_i)(1-\exp(-\sigma(\vx_i)\delta(\vx_i)))$, where $T$ denotes transmittance, $\sigma$ denotes density.

We modify the volume rendering equation to reflect our background design, a single point with full density:
\begin{equation}
  \vf(\vr) = \sum^{\Nfg}_{i=1}T_{i}(1-\exp(-\sigma_i\delta_i))\rfg_i + T^{\bg}\rbg,
  \label{eq:volume}
\end{equation}
where $\vf(\vr)$ is an aggregated pixel feature along the ray $\vr$, $T_i=\exp(\sum^{i-1}_{j=1}-\sigma_j\delta_j))$ denotes accumulated transmittance at $i$-th point $\vx_i$, $\mathbf{\Phi}_{i}$ and $\sigma_{i}$ are the feature and the density at $\vx_i$, and $\delta_{i}=t_{i+1}-t_{i}$ denotes the distance between adjacent points.
Since the background point is considered opaque and proceeded by all foreground points, we define its contribution using only the transmittance $T^{\bg}=\exp(\sum^{\Nfg}_{j=1}-\sigma_j\delta_j))$.
It is equivalent to placing an opaque background behind the scene in computer graphics techniques.

To synthesize high-resolution images,
we employ a 2D-CNN-based super-resolution module to upsample and refine the feature maps to an RGB image as commonly done in recent methods~\cite{niemeyer2021giraffe,xue2022giraffe,gu2021stylenerf,chan2022efficient}.

\subsection{Training objectives}
\label{sec:losses}

We use the non-saturating GAN loss $\Ladv$~\cite{goodfellow2014generative} and R1 regularization $\Lr$~\cite{mescheder2018training}.
Additionally, we use two regularizations.

\myparagraph{Background transmittance loss}
To ensure clear separation between foreground and background, we introduce new regularization on $T^{bg}$. 
The ray through the foreground region in the image should have a high foreground density that makes $T^{bg}$ close to 0, and thus the background feature should not affect the aggregated pixel.
In contrast, foreground density should be small enough to make $T^{bg}$ close to 1 when the ray corresponds to the background, so the aggregated pixel feature should be the same as the background feature.
Therefore, we induce the transmittance of the background to be binarized:

\begin{equation}
\mathcal{L}_{\bg} = \sum \min (T^{\bg}, 1-T^{\bg}).
\label{bgreg}
\end{equation}

\myparagraph{Foreground density loss}
To encourage clear shape, we use foreground regularization to prevent foreground density from diffusing.
Similar to Mip-NeRF 360\cite{barron2022mip}, our foreground loss penalizes the entropy of the aggregation weights on the ray to locate foreground points in the area where the actual geometry is located:
\begin{equation}
\mathcal{L}_{\fg} = \sum_{r} \left ( \sum_{i,j} \mathbf{w}_i^r \mathbf{w}_j^r |t_i^r - t_j^r | + \frac{1}{3} \sum_{i} {\mathbf{w}_{i}^{r}}^2 \delta_i^r \right ),
\label{eq:fgreg}
\end{equation}
where $i$ and $j$ are the indices of the weight, $r$ is the index of the ray, $\delta_{i}=t_{i+1}-t_{i}$ is the distance between adjacent points and $\mathbf{w}$ is the aggregation weights after sigmoid function.
This regularization is the integral of the weighted distance between all pairs of points on each ray.

The total loss function is then

\begin{equation}
    \mathcal{L}_{\text{total}} = \Ladv + \lambda_{\text{R}_{1}}\Lr + \lambda_{\fg}\mathcal{L}_{\fg} + \lambda_{\bg}\mathcal{L}_{\bg},
\end{equation}
where $\lambda_{\text{R}_{1}}, \lambda_{\fg}$ and $\lambda_{\bg}$ are hyperparameters.

\begin{figure}
\begin{center}
    \includegraphics[width = \linewidth]{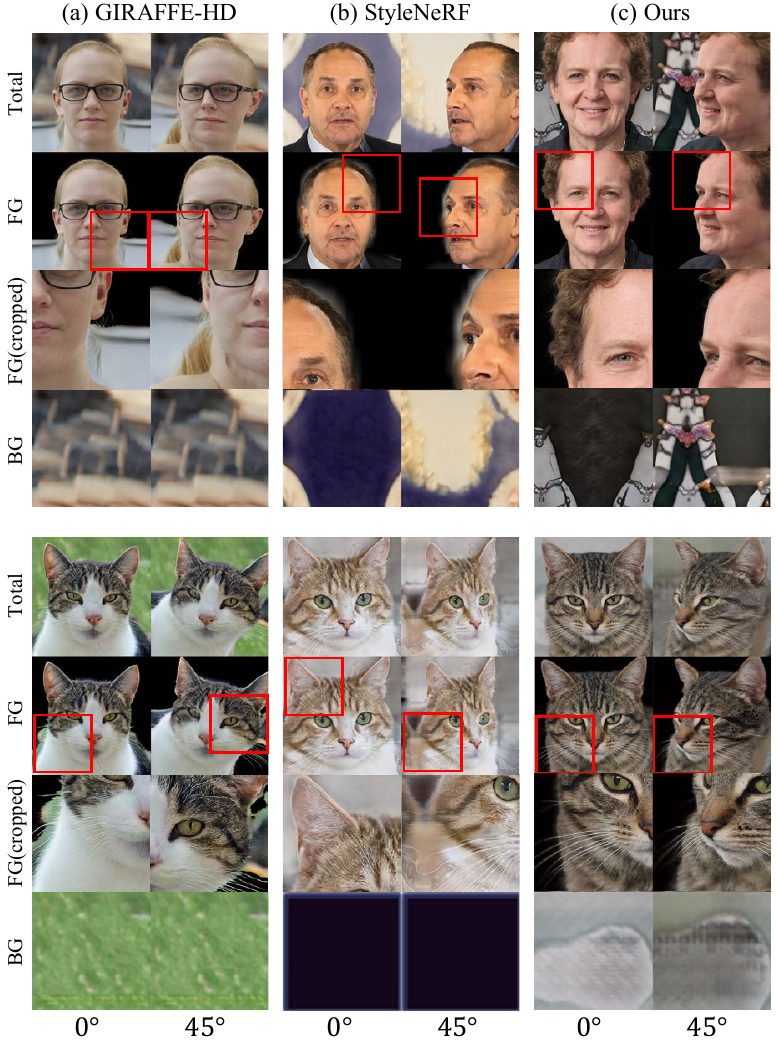}
    \vspace{-6mm}
    \caption{\textbf{Separate renderings of the foreground and background.} For easy comparison, we also show cropped foreground images.}
    \vspace{-3mm}
    \label{fig:separation}
\end{center}
\end{figure}

\section{Experiments}
\label{sec:exp}

In this section, we evaluate the effectiveness of BallGAN compared to the baselines regarding the faithfulness of foreground-background separation in \Sref{exp:separation}, effectiveness on complex backgrounds in \Sref{exp:effect}, the faithfulness of underlying 3D geometry in \Sref{exp:geometry}, and image quality in \Sref{exp:imagequality}. Implementation details are in \Aref{supp:implementation detail}.

\myparagraph{Datasets}
We validate our method on two front-facing datasets, FFHQ~\cite{karras2019style} and AFHQv2-Cats~\cite{Choi_2020_CVPR,karras2021alias}, and one 360$\degree$ dataset, CompCars~\cite{yang2015large}.
FFHQ has 70K images of real human faces, and AFHQv2-Cats contains 5,558 images of cat faces. We resize the resolutions of these datasets to $512^2$.
CompCars contains 136K images of cars with various resolutions and aspect ratios. In CompCars, we use a center cropping for each image and resize it to $256^2$.

\myparagraph{Competitors}
For our main comparisons we use EG3D~\cite{chan2022efficient}, StyleNeRF~\cite{gu2021stylenerf} and GIRAFFE-HD~\cite{xue2022giraffe}. We include EpiGRAFF~\cite{skorokhodov2022epigraf}\footnote{By incorporating NeRF++'s inverse sphere parameterization, EpiGRAF can separate foreground and background, same as StyleNeRF. However, the reported performance in the paper is based on a setting without the utilization of background representation. The official repository indicates a performance drop of approximately 10\% to 15\% when background representation is employed. Therefore, we employ the official version of EpiGRAF that doesn't use the background representation as a competitor. Refer to the \Aref{supp:qual compare} for a detailed ablation study using EpiGRAF, which adopts NeRF++ as the background representation.}, MVCGAN ~\cite{zhang2022multi}, VolumeGAN~\cite{xu2021volumegan} and StyleSDF~\cite{orel2022stylesdf} for quantitative comparisons.

\subsection{Foreground separation}
\label{exp:separation}
To achieve reasonable 3D perception and applicability, accurately separating foreground and background is an important evaluation factor.
As the background on a spherical surface is one of the key components of our method, we evaluate the separability and geometry of foregrounds against GIRAFFE-HD and StyleNeRF.
EG3D is excluded because it does not provide separation. 

\myparagraph{Comparison}
\Fref{fig:separation} shows rendered images of foreground and background, respectively.
GIRAFFE-HD uses an alpha mask for detailed foreground separation, but it relies on 2D feature maps instead of understanding the 3D scene.
Therefore, the foreground partly includes the background.
StyleNeRF shows some ability to separate the foreground on FFHQ, but fails to do so for all cases of AFHQ-cats, which contain a significant amount of fine-grained details.
By contrast, our results demonstrate fine-grained foreground separation, including intricate details like cat whiskers. Please refer to \Aref{supp:user} for quantitative evaluation (User study).

\begin{figure}
\begin{center}
    \includegraphics[width = \linewidth]{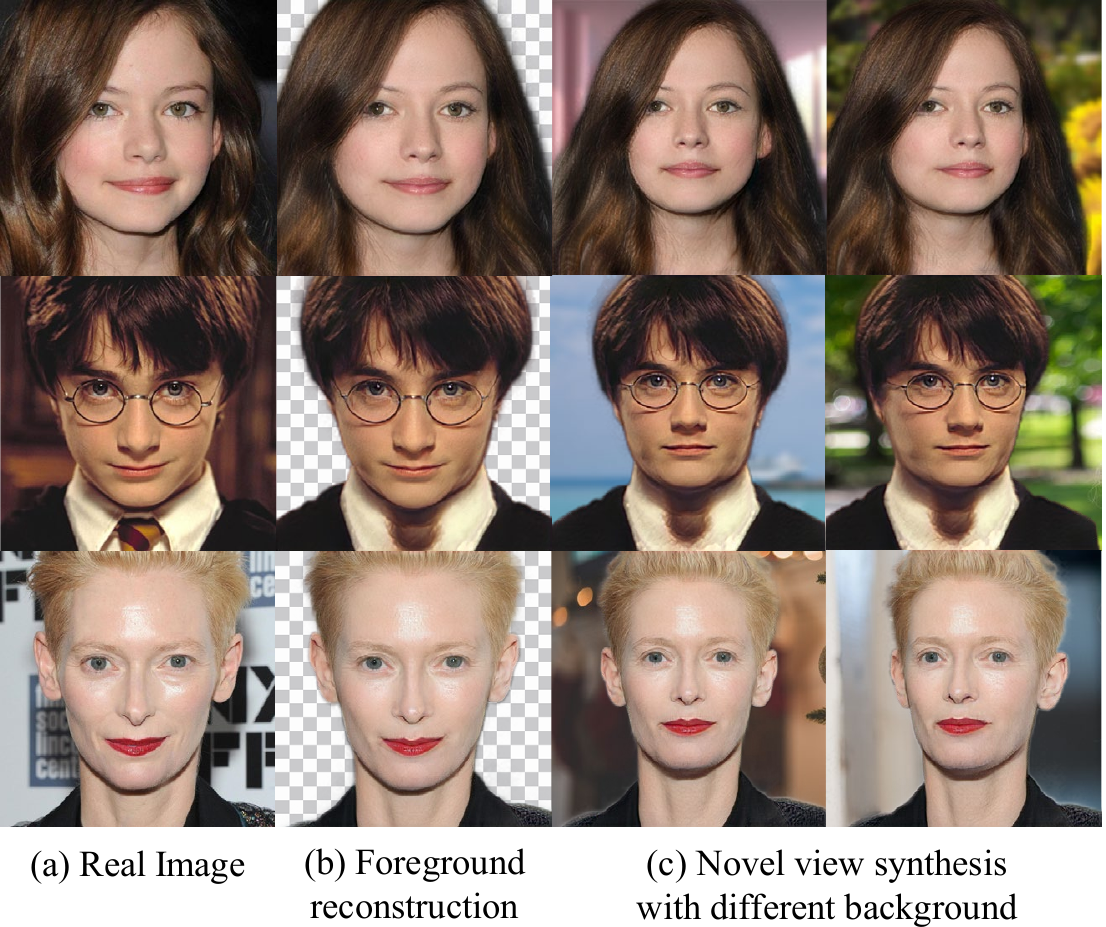}
    \vspace{-6mm}
    \caption{\textbf{Compositing foreground in different viewpoints on arbitrary backgrounds.} (a) is a target image, and (b) is a reconstructed foreground of ours using PTI)~\cite{roich2022pivotal}. (c) is a result of novel views on arbitrary backgrounds. By changing the camera pose and FOV, we show that our model can generate attributes of unobserved regions well.}
    \label{fig:app}
    \vspace{-3mm}
\end{center}
\end{figure}

\begin{figure}
\begin{center}
    \centering
    \includegraphics[width=\linewidth]{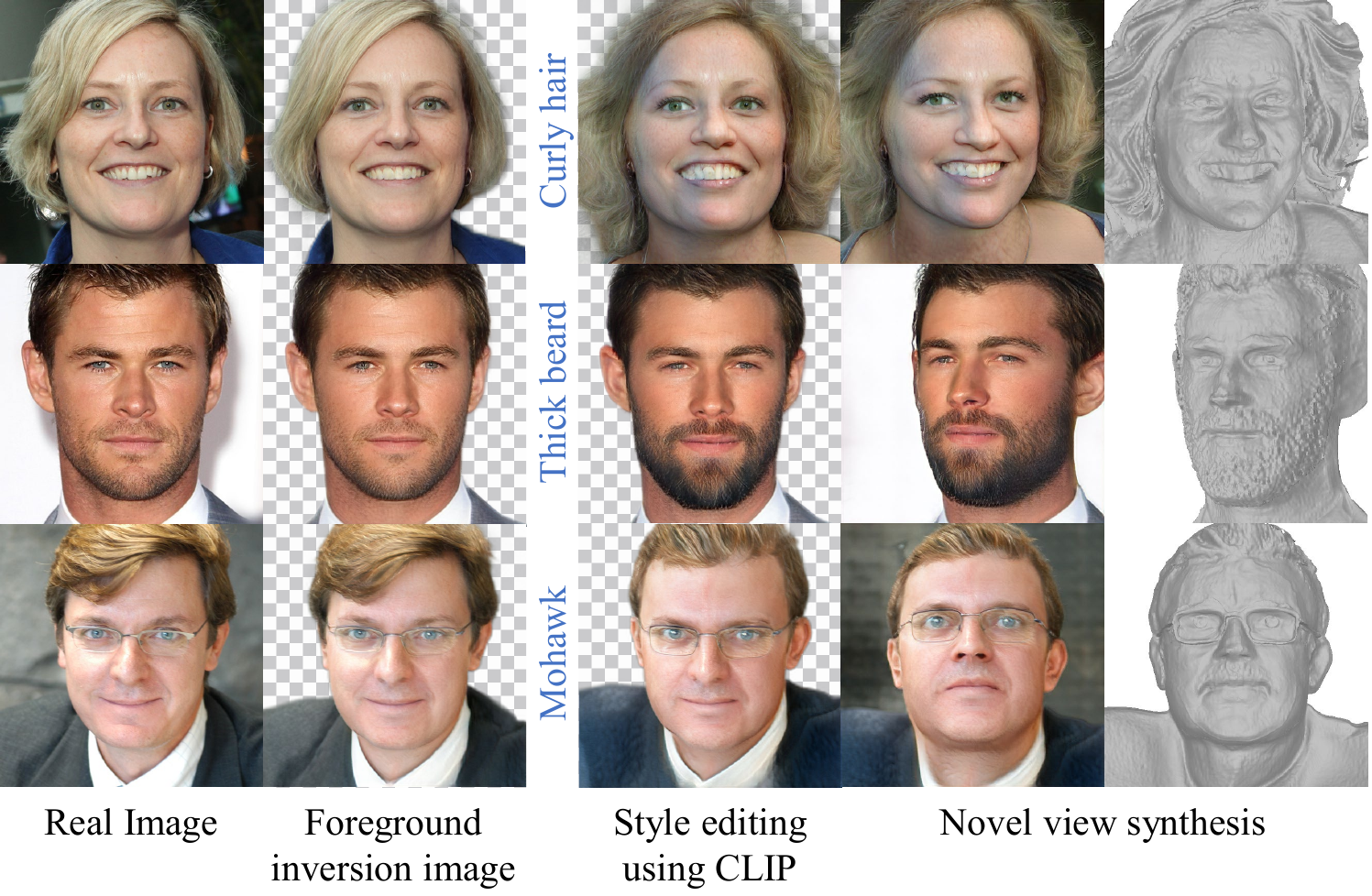}
    \vspace{-6mm}
    \caption{\textbf{CLIP guided editing results.} Given text prompt is \textcolor{blue}{blue}.}
    \label{fig:styleclip}
    \vspace{-3mm}
\end{center}
\end{figure}

\myparagraph{Content creation}
\Fref{fig:app} demonstrates the content creation capabilities achievable with BallGAN. 
Given a real image, its inversion on BallGAN provides 3D foreground that can be rendered in novel views and combined with different backgrounds. 
The alpha channel for the background is computed from the background transmittance in the volume rendering step, i.e., the last term in \eqref{eq:volume}. Even the facial regions that are not seen in the original images are realistic in the rendered images, such as parts of hair or chin. 
Note that \Fref{fig:app} has a wider field-of-view than the standard to produce more diverse results. 

\Fref{fig:styleclip} demonstrates the potential of BallGAN to 3D content creation. 
We can synthesize novel views of the edited foregrounds by inverting images to the latent space and using text-guided latent editing~\cite{patashnik2021styleclip}. 
Note that the 3D shapes are properly changed by the editing, e.g., hair. 
Therefore, BallGAN is useful for 3D content creation thanks to its foreground-background separation.

\subsection{Effectiveness on complex backgrounds}
\label{exp:effect}
Here, we demonstrate the effectiveness of our idea on complex backgrounds and wide camera angles, \ie CompCars dataset.
To use CompCars dataset where EG3D is not applicable due to the absence of a camera pose estimator, we apply a sphere background to StyleNeRF, namely \textbf{\textit{BallGAN-S}}.

\begin{figure}
\begin{center}
    \includegraphics[width =0.85\linewidth]{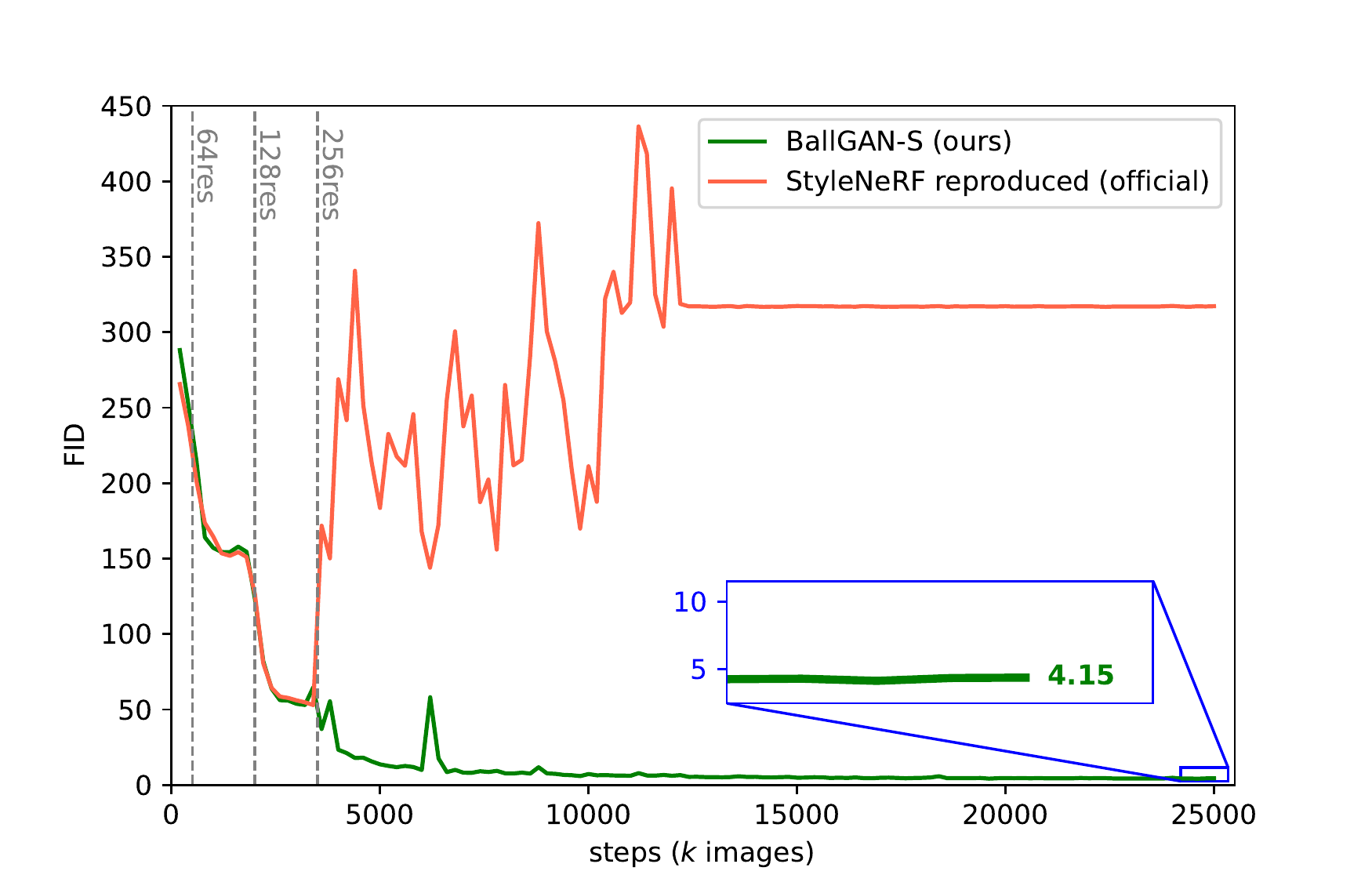}
    \vspace{-3mm}
    \caption{\textbf{FID over iterations on CompCars $\mathbf{{256^2}}$.} The FID score of StyleNeRF increases at ${256^2}$ and becomes constant around 12K steps. In contrast, BallGAN-S exhibits stable training and achieves notably low FID score.} 
    \label{fig:FID graph}
    \vspace{-3mm}
\end{center}
\end{figure}

\myparagraph{Training stability}
\Fref{fig:FID graph} compares image quality of BallGAN-S and StyleNeRF using Fr\'echet Inception Distance (FID)~\cite{heusel2017gans} over iterations. While StyleNeRF diverges as the image resolution grows from $128^2$ to $256^2$\footnote[2]{This phenomenon is also reported in the official repository.}, BallGAN-S smoothly converges below the reported FID of StyleNeRF. It implies that our method is generally beneficial to different foreground backbones and greatly improves training stability. 

\begin{figure}
  \centering
  \begin{subfigure}{\linewidth}
  \centering
    \includegraphics[width =\linewidth ]{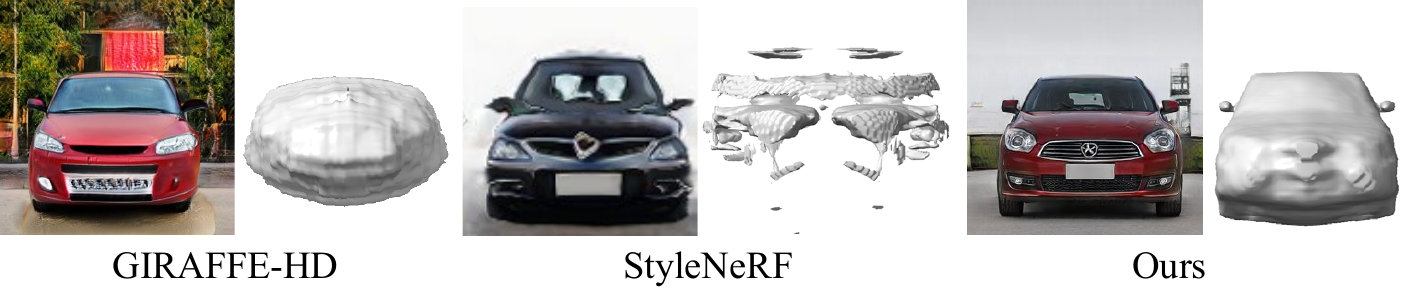}
    \vspace{-6mm}
    \caption{Qualitative comparison of generated images and their  corresponding 3D geometry.}
    \label{fig:BallGAN-S_a}
  \end{subfigure}
  \hfill
  \begin{subfigure}{\linewidth}
  \centering
    \includegraphics[width = \linewidth ]{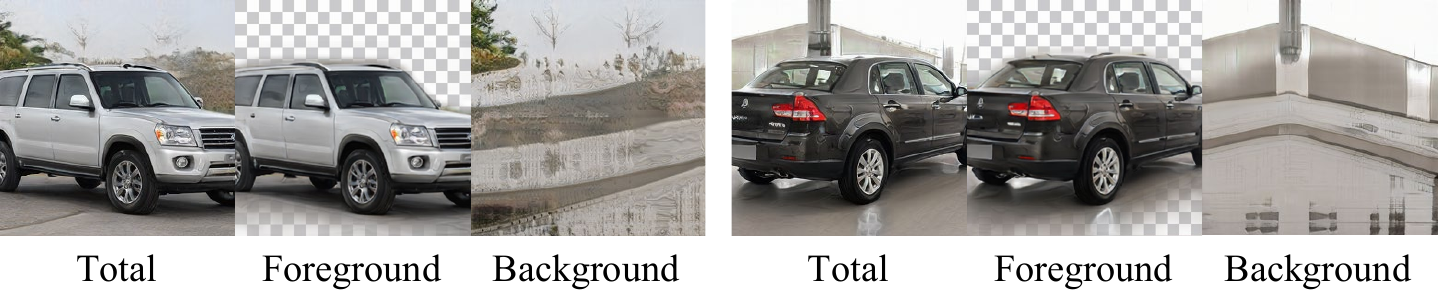}
    \vspace{-6mm}
    \caption{Separate renderings with BallGAN-S}
    \label{fig:BallGAN-S_b}
  \end{subfigure}
  \vspace{-5mm}
  \caption{\textbf{Results of BallGAN-S on CompCars $256^2$.}}
  \label{fig:BallGAN-S}
  \vspace{-2mm}
\end{figure}

\myparagraph{Comparisons}
In \fref{fig:BallGAN-S}, we present qualitative results of BallGAN-S, which showcase the robustness of our design on CompCars. 
\fref{fig:BallGAN-S_a} shows that both GIRAFFE-HD and StyleNeRF exhibit a deficiency in fidelity in their modeled 3D compared to the quality of the generated images. 
On the other hand, ours maintains a high level of fidelity for both images and 3D models.
In \fref{fig:BallGAN-S_b}, we demonstrate that our simple yet effective idea ensures successful separation of foreground and background, even for datasets with complex backgrounds and wide camera angles. 
Quantitative comparisons will be addressed in \Sref{exp:imagequality}


\subsection{Faithfulness of the underlying 3D geometry}
\label{exp:geometry}

It is essential for 3D-aware GANs to model the correct 3D geometry of the scenes so that their rendered images on arbitrary camera poses are convincing views of the real 3D scenes. Quantitative comparisons are followed by qualitative comparisons.

\input{tables/3D.tex}

\input{tables/colmap}

\myparagraph{Quantitative results}
We quantitatively compare the underlying 3D model following the protocols in EG3D~\cite{chan2022efficient}. In \Tref{tab:3D}, ID measures multi-view facial identity consistency\footnote[3]{The mean Arcface~\cite{deng2019arcface} cosine similarity}, Depth indicates MSE of the expected depth maps from density against estimated depth-maps\footnote[4]{Estimations for Depth and Pose are from \cite{deng2019accurate}} in frontal view, and Pose implies controllability by MSE between the estimated pose of synthesized image and the input (target) pose. \Aref{supp:eval} describes further details of the protocol. 
BallGAN outperforms the baselines in all metrics evaluating 3D geometry.

We further push the evaluation: the number of reconstructed points from 128 views by COLMAP~\cite{schonberger2016structure} in five inverted samples of FFHQ training set. \Tref{tab:colmap} provides the numbers and example point clouds of the methods. Since COLMAP reconstructs the points with high photometric consistency, the larger number of points indicates higher multi-view consistency. 
BallGAN demonstrates superior performance in terms of multi-view consistency, especially in the face and hair region where the number of reconstructed points is substantially higher than other methods.
While EG3D also achieves a similar number of reconstructed points as BallGAN, a large portion of these points lies on the background walls rather than the face.
As the comparison results show, our sphere background induces the synthesis of accurate foreground geometry, thereby improving multi-view consistency.

\begin{figure}[t]
\begin{center}
    \includegraphics[width = \linewidth]{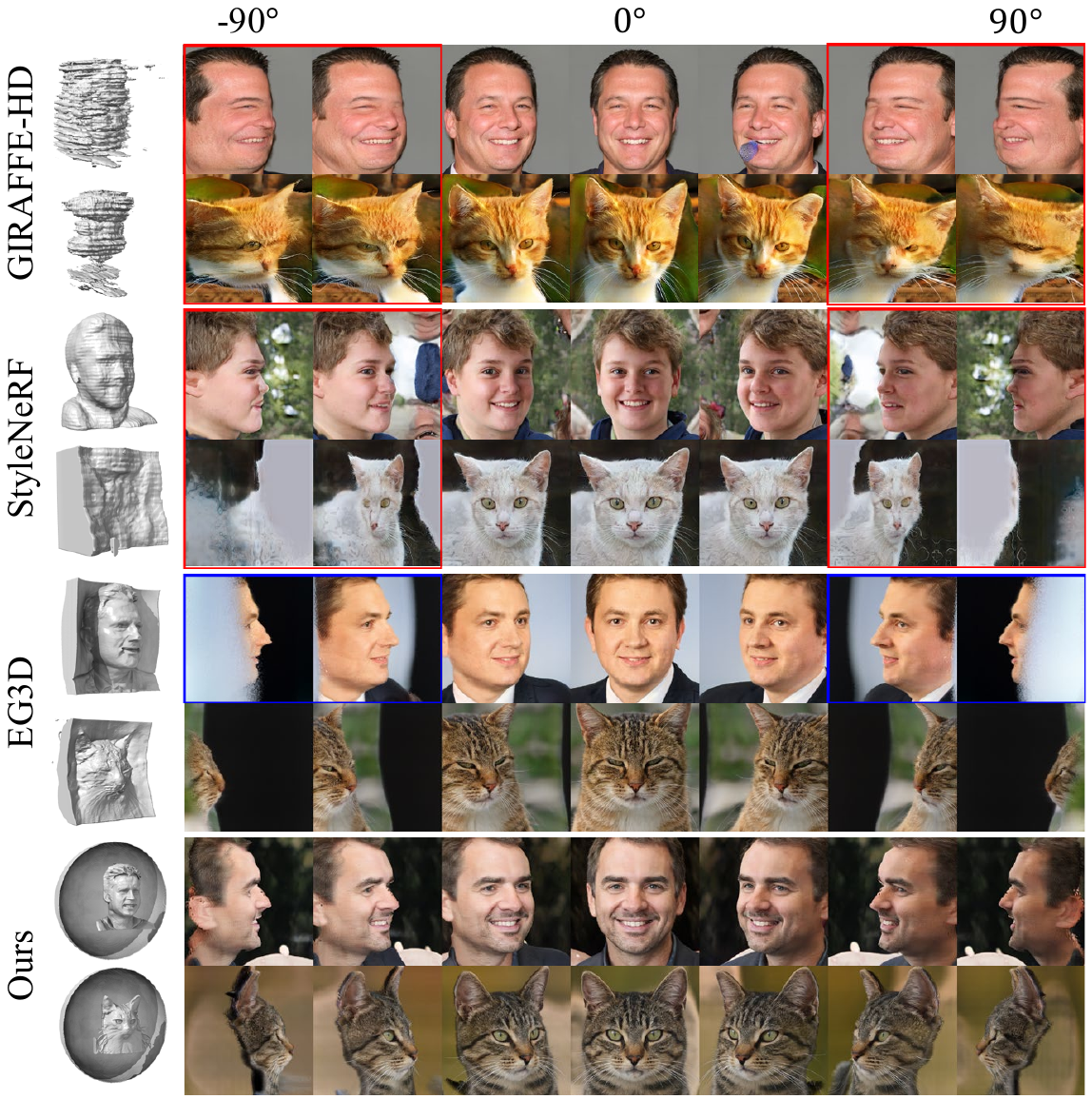}
    \vspace{-4mm}
    \caption{\textbf{Images rendered on various camera poses}. GIRAFFE-HD and StyleNeRF show distortions, especially on extreme camera poses (\textcolor{red}{red} boxes). The rendered images of EG3D are distorted by concave walls on extreme camera poses (\textcolor{blue}{blue} boxes). In contrast, BallGAN synthesizes realistic and multi-view consistent images.}
    \label{fig:varyingview}
    \vspace{-4mm}
\end{center}
\end{figure}

\myparagraph{Qualitative comparison: generated scenes} \Fref{fig:varyingview} compares how each method renders \emph{generated} scenes on different perspectives, expecting the images to have multi-view consistency and realism. The leftmost column provides meshes of the scene for reference.
We notice severe distortions in GIRAFFE-HD and StyleNeRF when the camera rotates more than $\pm60\degree$ implying their spurious 3D geometry (red box in \Fref{fig:varyingview}). 
This problem is evident in the marching cube results of GIRAFFE-HD, which separately models foreground and background but without their separate ranges.
StyleNeRF produces rough geometry and camouflages detailed shapes with color. 
Discussion on the missing backgrounds is deferred to \Aref{supp:qual compare}. 

Similarly, the rendered images of EG3D show distortions from $\pm60\degree$ angles, \eg, the ears are truncated first and then the cheeks at $\pm90\degree$ angles (blue box in \Fref{fig:varyingview}).
The mesh explains that the faces are engraved to a concave wall expanding from the ridge of the faces.
Furthermore, although the meshes show greater detail compared to StyleNeRF, there are areas of disagreement between the underlying geometry and its rendered images, \eg, the boundary between hair and forehead is fuzzy in the geometry, whereas it becomes clear after color rendering.

On the other hand, BallGAN synthesizes realistic images that maintain consistency across multiple views, even when rendered in extreme side views.
It implies that the separate background on a sphere removes the depth ambiguity and does not interfere with the foreground object. 
Notably, we observe a significant enhancement in fine details, such as hair and whiskers.
For a more detailed multi-view comparison with all baseline models, please refer to \Aref{supp:comparison}.

\begin{figure*}[t]
\begin{center}
    \includegraphics[width = \linewidth ]{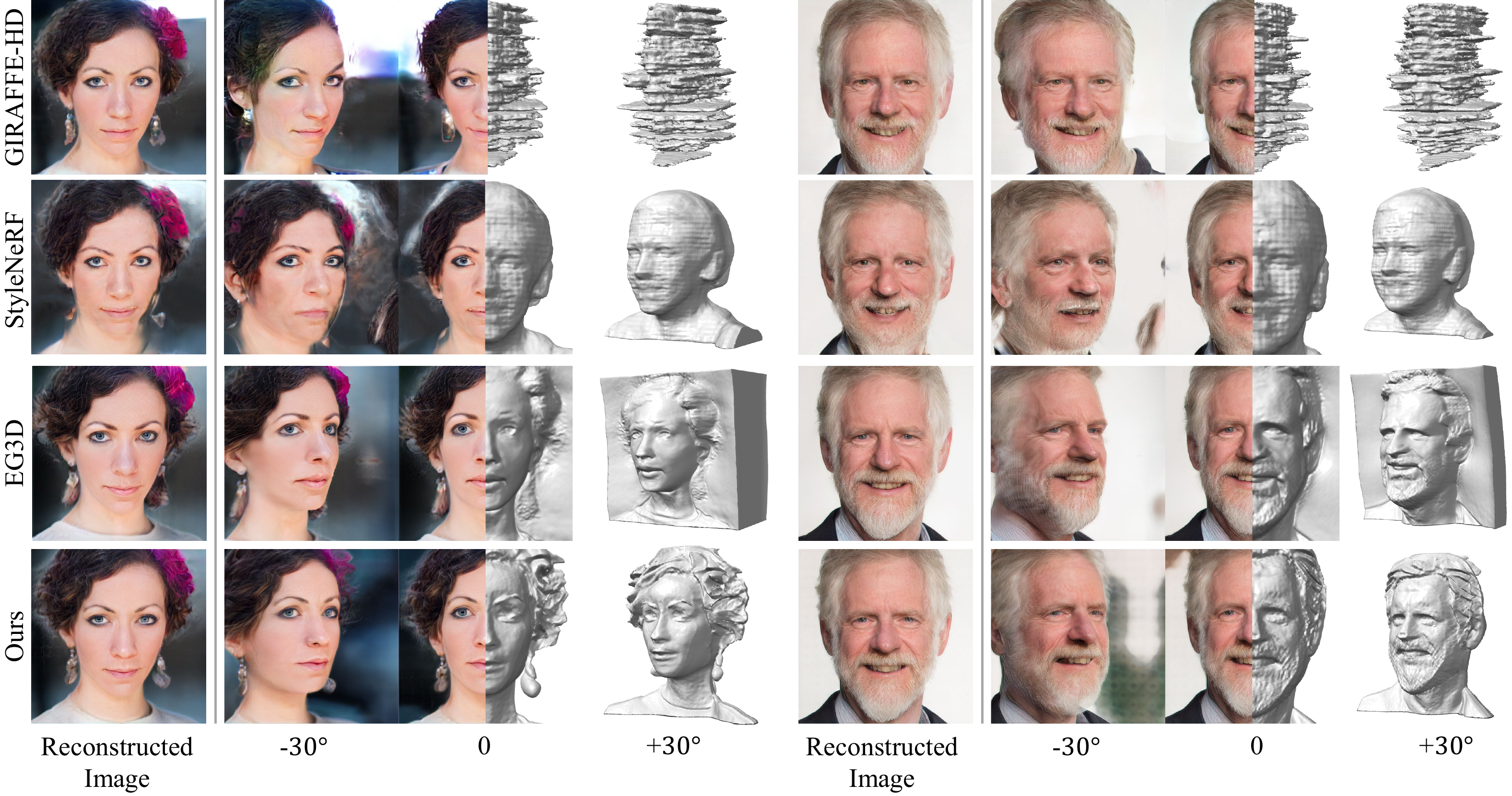}
    \vspace{-2em}
    \caption{\textbf{Renderings and marching cubes of the same samples.} Given real image omitted as all models faithfully reconstruct it. Although all methods render the target image close by inversion, the underlying 3D geometries of previous methods are all different. We adjusted the threshold for each mesh at the line where the pupils do not break.}
    \label{fig:geometry}
    \vspace{-4mm}
\end{center}
\end{figure*}
\myparagraph{Qualitative comparison: inversion of real images}
\Fref{fig:geometry} compares renderings and meshes of the same scenes through pivotal tuning inversion (PTI)~\cite{roich2022pivotal} of \emph{real} images from the training set. Although the image reconstructions of all methods are similar in target pose, the differences become more visible in different viewpoints and in their underlying 3D geometries. GIRAFFE-HD apparently produces geometry that least fits the rendered image and thus renders inconsistent images in different views. 
StyleNeRF captures only rough outlines and placements in the geometry so that color makes the rendered scene realistic. Especially, the mesh does not reveal the beard and the boundary between hair and forehead.
While EG3D can recover realistic geometry that mostly fits the given image, it has limitations such as faces being stuck to a wall. Moreover, it fails to accurately represent details such as eyebrows or accessories, which are evident in the input image. In contrast, BallGAN excels at accurately modeling the foreground in 3D space, and even faithfully represents the details shown in the images, such as wavy hair, earrings, and eyebrows.


\input{tables/imagequality.tex}

\begin{figure}[t]
\begin{center}
    \includegraphics[width = 0.95\linewidth]{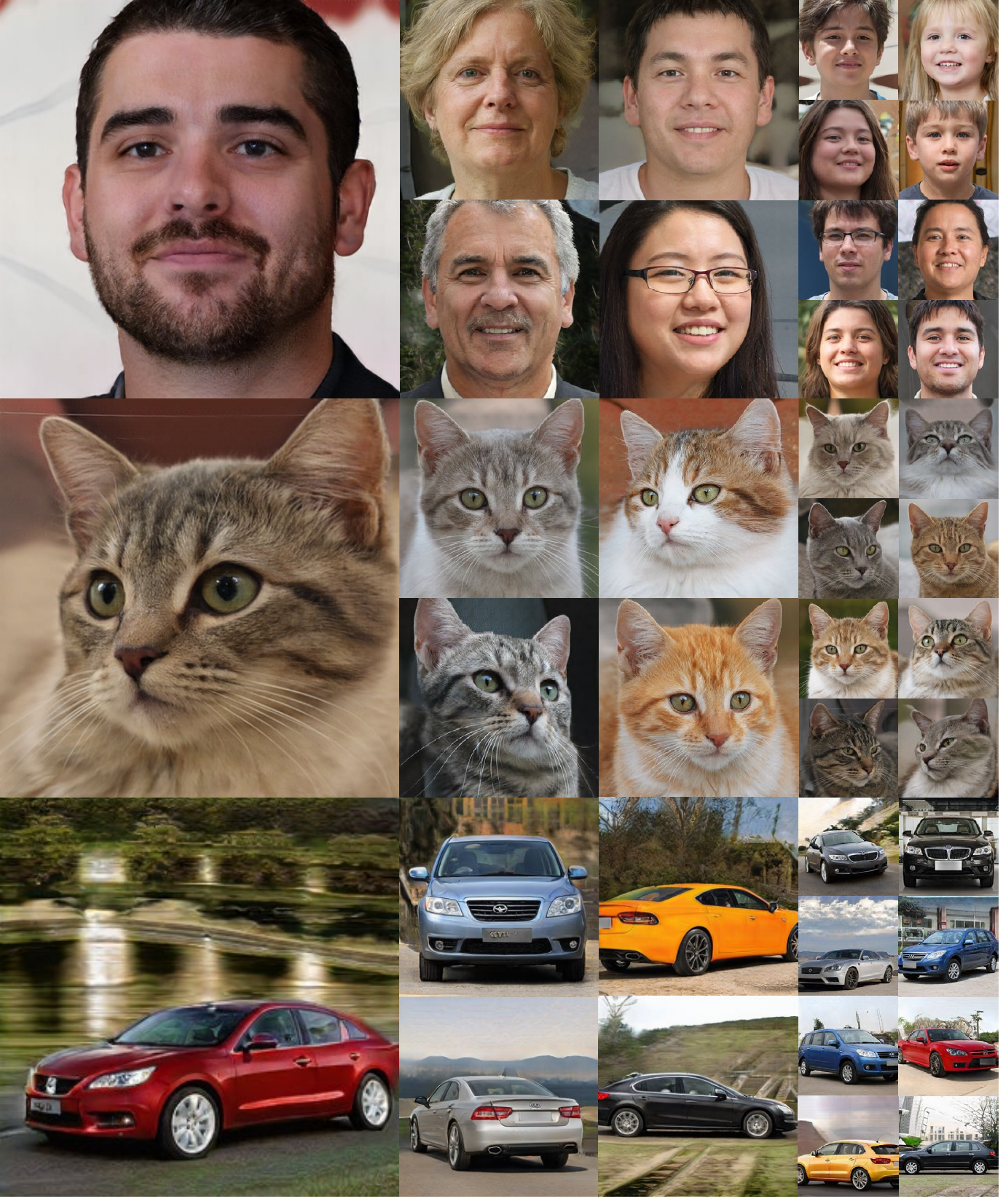}
    \vspace{-2mm}
    \caption{\textbf{Set of images generated by BallGAN.} We sample images of ${512^2}$ resolution from BallGAN on FFHQ ${512^2}$ and AFHQv2-Cats ${512^2}$, as well as ${256^2}$ resolution images from BallGAN-S on CompCars ${256^2}$. Each image is rendered with randomly sampled camera pose.}
    \label{fig:qual}
    \vspace{-6.5mm}
\end{center}
\end{figure}

\subsection{Image quality}
\label{exp:imagequality}

We evaluate generated image quality on the FFHQ $512^2$, AFHQv2-Cats $512^2$, CompCars $256^2$ datasets. Images for FFHQ $512^2$, AFHQv2-Cats $512^2$ are generated by BallGAN and images for CompCars $256^2$ are generated by BallGAN-S.

\myparagraph{Quantitative results}
\Tref{tab:quals} compares image quality in FID.
For FFHQ, AFHQv2-Cats, BallGAN outperforms all the baselines except EG3D.
Although EG3D achieves the best FID, it does not support foreground-background separation and suffers in generating 3D geometry (\Sref{exp:geometry}). Furthermore, EG3D requires camera poses of real images, which are not always available, e.g., CompCars. On the other hand, we achieve the state-of-the-art FID on CompCars with BallGAN-S and the second-best FID on FFHQ and AFHQv2-Cats closely following EG3D. We note that CompCars has more complex backgrounds and 360$\degree$ camera poses.

\myparagraph{Qualitative results}
\Fref{fig:qual} provides example images generated by BallGAN and BallGAN-S. Our models faithfully generate diverse samples in multiple views. More examples can be found in \Aref{supp:uncuratedimages}.


\begin{figure}[t]
  \centering
  \begin{subfigure}{\linewidth}
  \centering
    \includegraphics[width = 0.7\linewidth ]{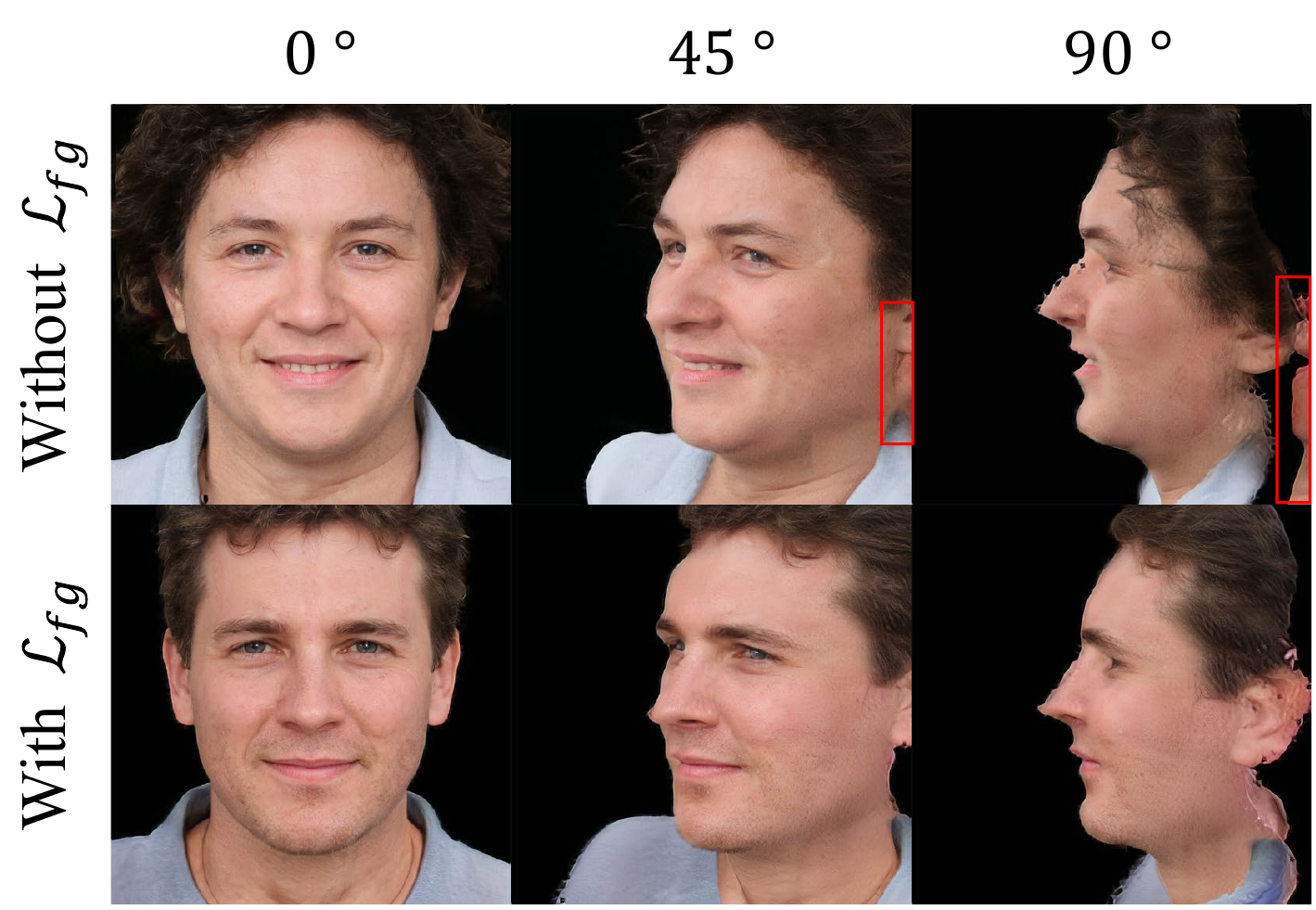}
    \vspace{-1mm}
    \caption{\textbf{Visual comparison on the effect of foreground density regularization.} Removing $\mathcal{L}_{\fg}$ introduces occasional floating objects behind the neck (red box).}
    \label{fig:fg}
    \vspace{-1mm}
  \end{subfigure}
  \hfill
  \begin{subfigure}{\linewidth}
  \centering
    \includegraphics[width = 0.85\linewidth ]{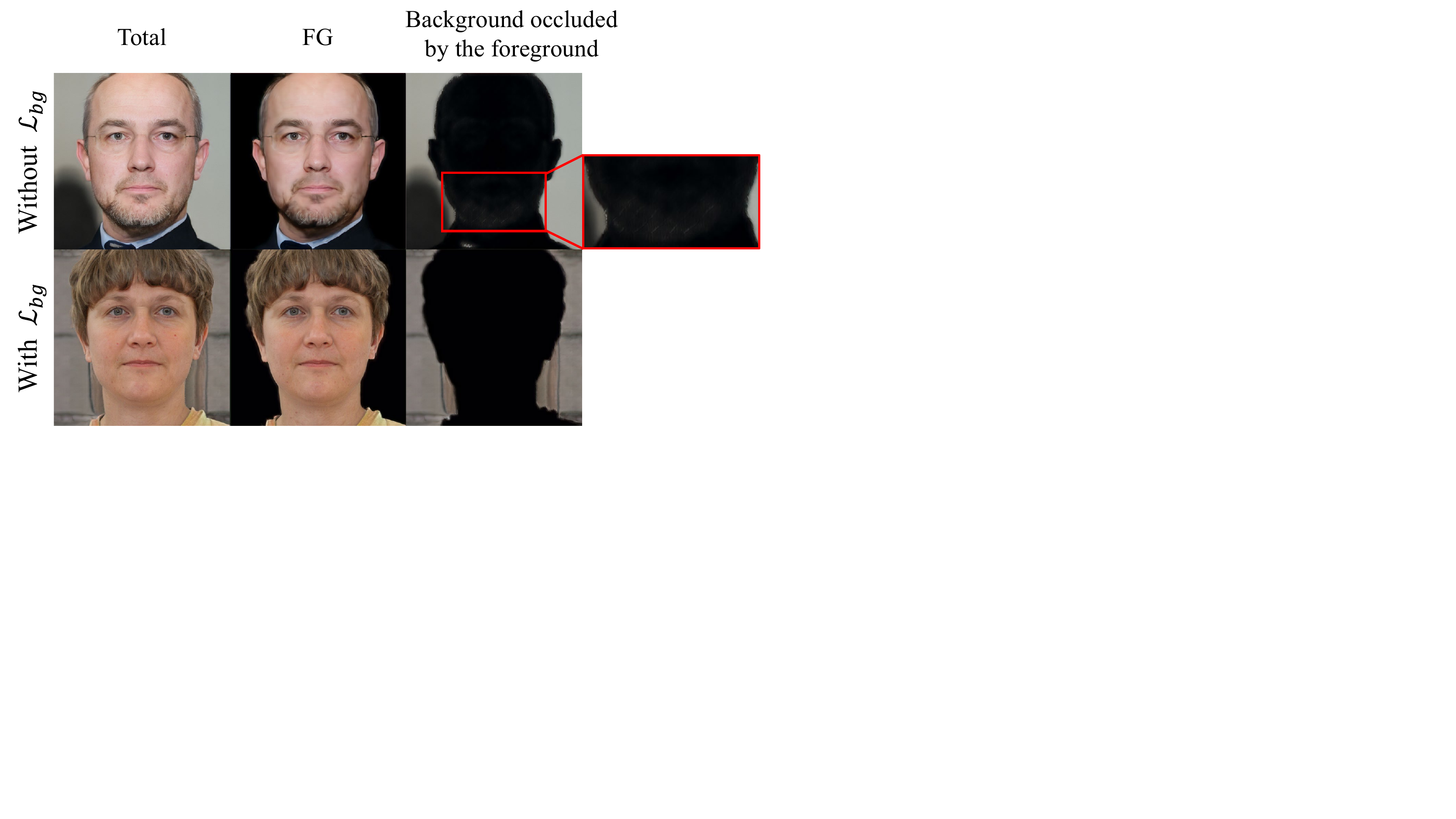}
    \vspace{-1mm}
    \caption{\textbf{Visual comparison on the effect of background transmittance regularization.} The use of $\mathcal{L}_{\fg}$ results in a completely opaque foreground, rendering the background occluded by the foreground as entirely black.}
    \label{fig:bg}
  \end{subfigure}
  \vspace{-6mm}
  \caption{\textbf{Ablations for two regularizations.}}
  \label{fig:abl}
  \vspace{-3mm}
\end{figure}

\subsection{Ablation of the losses}
\label{exp:losses}

We conduct ablation studies to evaluate the effect of the regularizers.
\Fref{fig:abl} shows the effects of our foreground and background regularization.
Without $\mathcal{L}_{\fg}$, BallGAN on FFHQ occasionally generates small floating objects behind faces. 
$\mathcal{L}_{\fg}$ mitigates scene diffusion, thus inhibiting the formation of subtle shape artifacts such as floating objects behind the object.
Additionally, using the background regularization $\mathcal{L}_{\bg}$, we get clearer foreground-background separation.
\Fref{fig:bg} shows that removing $\mathcal{L}_{\bg}$ allows the background to participate in synthesizing the foreground.
For the result without $\mathcal{L}_{\bg}$, the beard is not entirely black, indicating partial influence from the background (red box in \Fref{fig:bg}). 
In other words, the foreground is not fully opaque.
This is because the background transmittance loss $\mathcal{L}_{\bg}$ encourages the foreground density to either completely block or leave the space empty before the rays hit the background.

\section{Conclusion}

We propose a 3D-aware GAN framework named BallGAN, which represents a scene as a 3D volume within a spherical surface, enabling the background representation to lie on a 2D coordinate system.
This approach resolves the challenges of training a generator to learn a 3D scene from only 2D images.
Our proposed framework successfully separates the foreground in a 3D-aware manner, which enables useful applications such as rendering foregrounds from arbitrary viewpoints on top of given backgrounds.
BallGAN also achieves superior performance in 3D awareness, including multi-view consistency, pose accuracy, and depth reconstruction. 
Additionally, our approach shows significant improvement in capturing fine image details in 3D space, compared to existing methods.

\noindent\textbf{Acknowledgements} 
{\small This work was supported by the National Research Foundation of Korea(NRF) grant funded by the Korea government(MSIT) (No. 2022R1F1A1076241). The part of experiments was conducted on NAVER Smart Machine Learning (NSML) platform~\cite{kim2018nsml, sung2017nsml}.}

\clearpage

{\small
\bibliographystyle{ieee_fullname}
\bibliography{egbib.bib}
}

\clearpage
\title{Supplemental Material for BallGAN}

\maketitle
\appendix

\renewcommand{\thetable}{S\arabic{table}}
\renewcommand{\thefigure}{S\arabic{figure}}
\setcounter{figure}{0}
\setcounter{table}{0}

We provide the following supplementary materials:

\begin{enumerate}[label=\Alph*,nosep]
    \item Background design choice
    \item Effectiveness of background representation
    \item Ablation of the losses
    \item Implementation details
    \item User study
    \item Evaluation protocols
    \item Detailed qualitative comparison
    \item More comparison with EG3D
    \item Detailed multi-view comparison
    \item Uncurated samples
\end{enumerate}


\section{Background design choice}
\label{supp:bg design}

\begin{figure}[b]
    \centering
    \includegraphics[width=\linewidth]{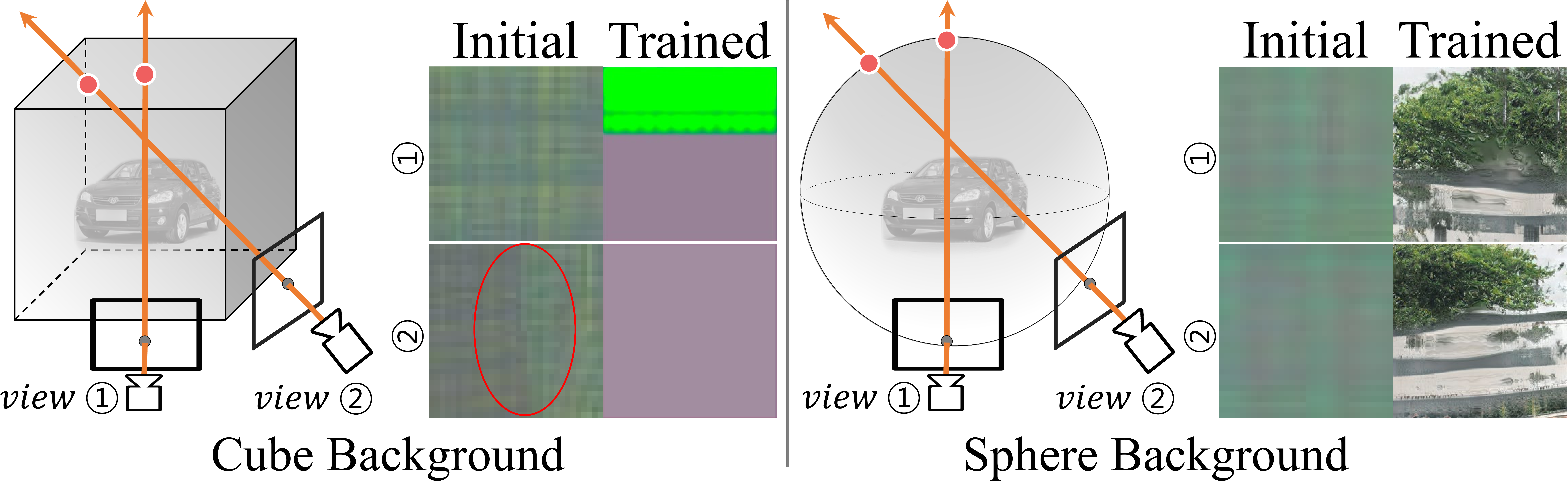}
    \caption{\textbf{Background should be modeled spherical rather than cubic.} While the edges of the cube are reflected in the rendered images (\textit{Initial}), the sphere has no such artifacts in the rendered images. While the cubic background fails to produce plausible images, our spherical background produces sensible backgrounds (\textit{Trained}). }
    \label{fig:background_design}
\end{figure}

This section explains the rationale why our background has a spherical shape rather than anything else.
Notably, our goal is not to accurately model the geometry of the background, but rather to ensure that the integrity of the foreground of interest is not compromised.
To ensure that the background is taken into consideration from all possible angles, it is imperative that the background encompasses the camera sphere. 
For instance, a planar background fails to cover the background when the camera rotates beyond $90\degree$ from its normal vector. 

Even if the view frustum can account for the entire background, any abrupt changes in gradient or inconsistencies in distances from the camera can engender unstable learning. 
To analyze the background effect, we trained BallGAN-S on the CompCars dataset with various complex background representations that occupy a significant portion of the image, using only different representations of the background such as sphere and cube, in \Fref{fig:background_design}. The cube background does not converge. 
Therefore, the sphere background is the only reasonable choice for background representation. 

\section{Effectiveness of background representation}
\label{supp:SSO}

\begin{figure}
\begin{center}
    \centering
    \includegraphics[width=\linewidth]{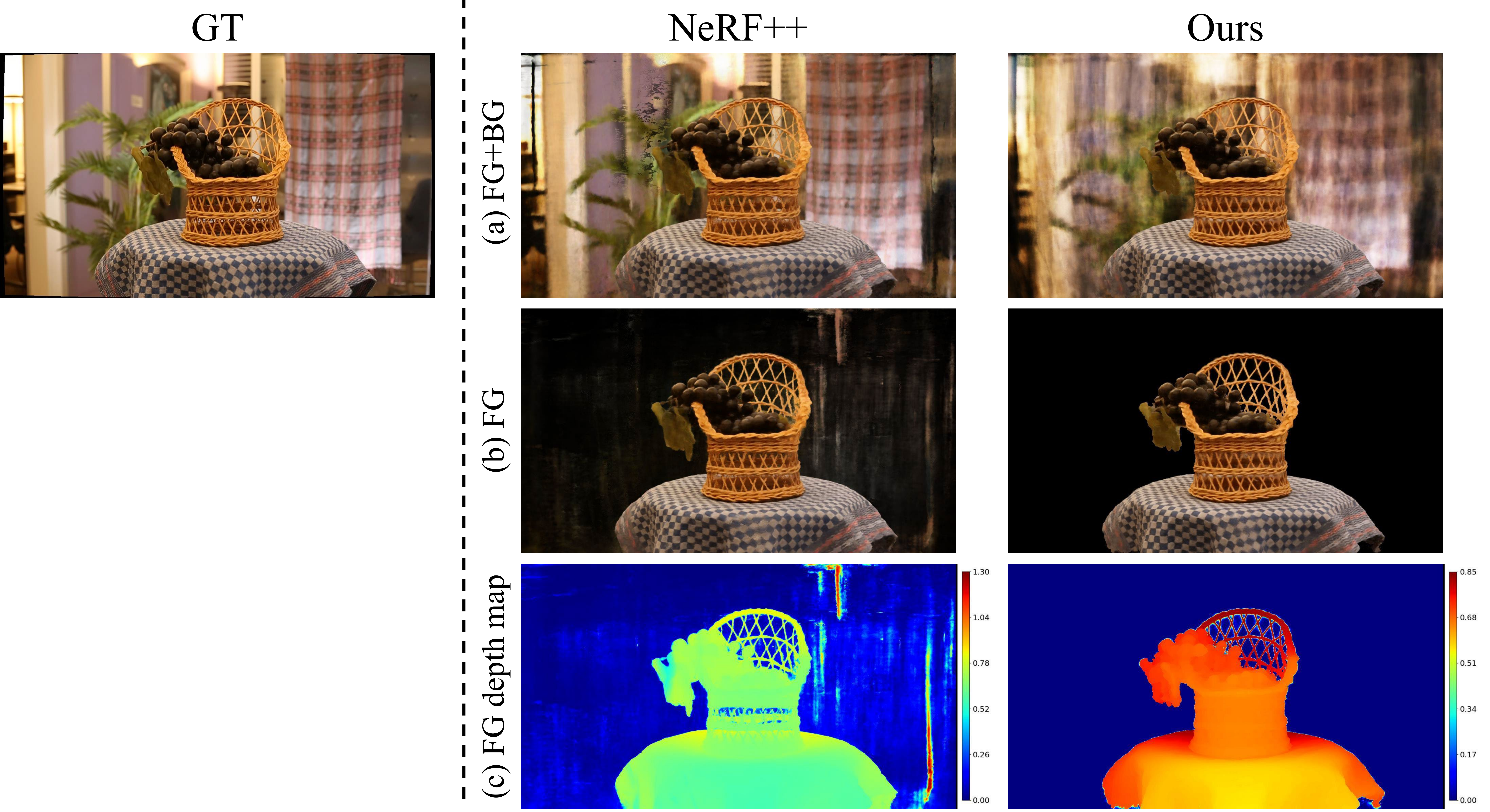}
    \caption{\textbf{Effectiveness of our spherical background on single scene overfitting scenario.} The sole foreground rendering and depth map demonstrates our spherical background is beneficial for capturing foreground geometry}
    \label{fig:sso}
\end{center}
\end{figure}

In this section, we demonstrate the effect of our spherical background representation, which enhances the focus on the foreground. 
We verify the efficacy of our background representation through a single-scene overfitting (SSO) experiment, in which we overfit a 3D model to a single scene captured by multi-view images, namely lf-basket~\cite{yucer2016efficient}. 
We use the vanilla NeRF~\cite{mildenhall2020nerf} for the foreground, and keep the spherical background representation. 
In other words, NeRF++ and Ours differ only in the background representation. 

As shown in \Fref{fig:sso}, NeRF++ does not clearly distinguish between foreground and background, and the estimated depth is erroneous, e.g., the table has a lower depth at the deepest end. 
In contrast, our approach clearly separates foreground and background and better estimates foreground depth. 
Thus, our design demonstrates effectiveness in focusing resources on learning foreground 3D geometry.

\input{tables/supp_abl.tex}

\section{Ablation of the losses}
\label{supp:losses}

We conduct ablation studies to evaluate the impact of each regularization on image quality. 
\Tref{tab:qual_abl} shows the effects of our foreground and background regularization.
Applying the foreground density loss $\mathcal{L}_{\fg}$ improves FID.
The background transmittance regularization $\mathcal{L}_{\bg}$ not only facilitates a clearer separation between foreground and background but also enhances FID score.


\section{Implementation details}
\label{supp:implementation detail}

\myparagraph{BallGAN}
Our implementation mostly follows the official implementation of EG3D\footnote{https://github.com/NVlabs/eg3d} including training hyperparameters, dual discrimination, pose-conditioning on discriminator, two-stage training, equalized learning rates~\cite{karras2018progressive}, a mini-batch standard deviation layer at the end of the discriminator~\cite{karras2018progressive}, exponential moving average of the generator weights, a non-saturating logistic loss~\cite{goodfellow2014generative}, and R1 regularization~\cite{mescheder2018training} with $\gamma=1$.
We also use the same camera intrinsic parameters and FFHQ preprocessing from EG3D.

\begin{figure}[h]
\begin{center}
    \centering
    \includegraphics[width=\linewidth]{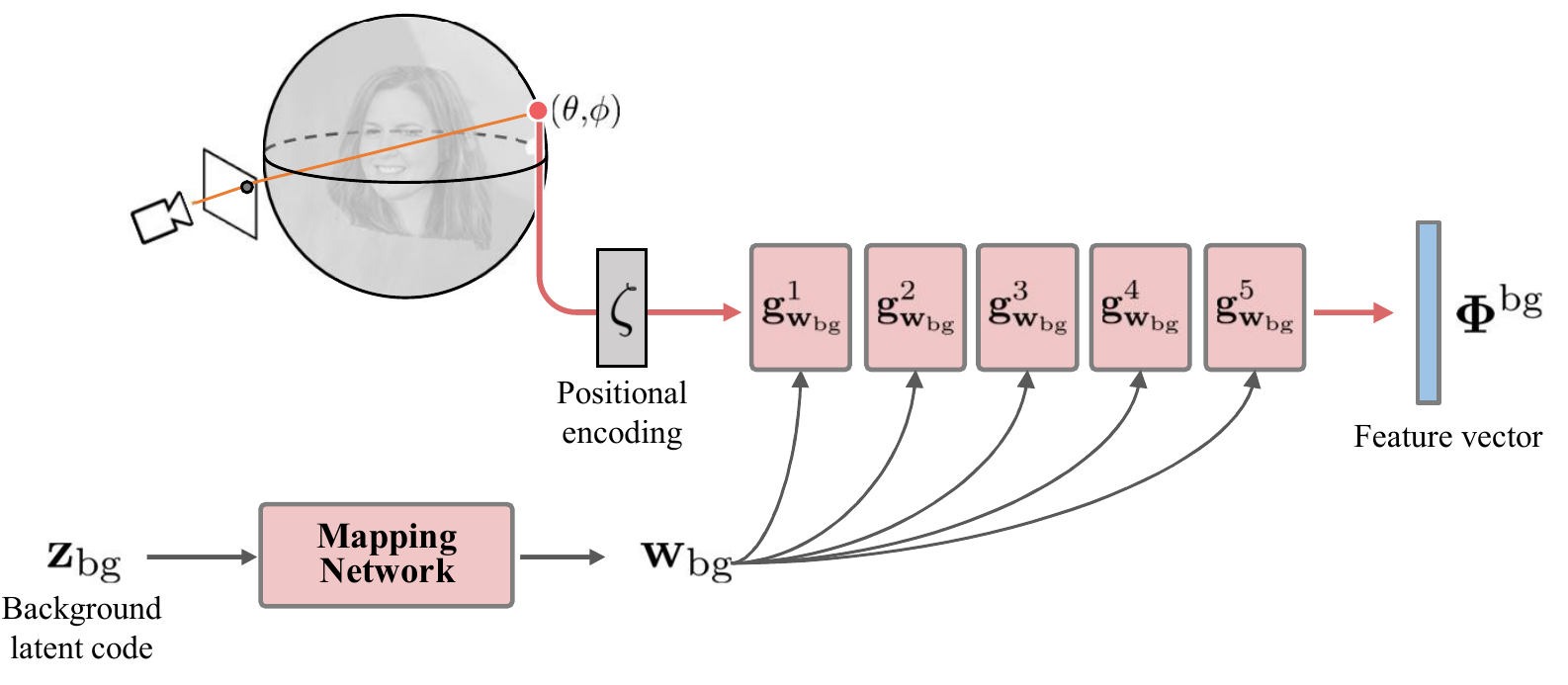}
    \caption{\textbf{Background architecture}}
    \label{fig:bgarch}
\end{center}
\end{figure}

The weights of the foreground density output layer are initialized to zero to guarantee the contribution of the background at the beginning of the training.
\Fref{fig:bgarch} illustrates the architecture for the background representation. A five-layer $1\times1$ convolutional network maps the positional encoding $\zeta$ of a background point to a feature vector. 
The style code from an eight-layer MLP, \ie, the mapping network, modulates the weights of the convolutions $\vg_{\wbg}$. 
The background representation mapping network shares the same design as the mapping network in StyleGAN2~\cite{Karras2019stylegan2}. The number of channels of the intermediate features are in \Tref{tab:bg_arc}.
The last layer has a sigmoid clamping from MipNeRF~\cite{barron2021mip} as in the foreground neural render of EG3D.
We use the positional encoding of $L=10$ on the background's 2D spherical coordinates. View direction is not considered for our background representation.

\input{tables/supp_bg_arc.tex}

On FFHQ, we schedule the coefficient of the foreground density loss $\lambda_\text{fg}$ to exponentially grow from 0 to 0.25 and the coefficient of the background transmittance regularization $\lambda_\text{bg}$ to exponentially grow from 0 to 1 in the first stage. We set the coefficients $\lambda_\text{fg}=1$ and $\lambda_\text{bg}=0.5$ in the second stage.

For AFHQv2-Cats, we start from the weights pretrained on FFHQ for the first step and fine-tune them on AFHQv2-Cats as done in EG3D. We set $\lambda_\text{fg}=\lambda_\text{bg}=0$ to let the foreground better capture the fine details such as whiskers.

\myparagraph{BallGAN-S}
BallGAN-S is a variant using StyleNeRF as a baseline instead of EG3D. We add the same background network on top of the official StyleNeRF implementation\footnote{https://github.com/facebookresearch/StyleNeRF}. We set $\lambda_\text{fg}=0.25$ and $\lambda_\text{bg}=0$.

\input{tables/supp_fid.tex}
\myparagraph{Competitors}
In the comparison experiments, we reported the best FIDs among the available sources: reported, official checkpoints, and official training code.
We used the official training codes as-is to reproduce FIDs if the official repository does not provide the checkpoints\footnote{https://github.com/genforce/volumegan}\footnote{https://github.com/universome/epigraf}\footnote{https://github.com/royorel/StyleSDF}\footnote{https://github.com/AustinXY/GIRAFFEHD}.

StyleNeRF, StyleSDF, EpiGRAF, and VolumeGAN do not provide training guidelines for AFHQv2-cats~\cite{Choi_2020_CVPR}.
For StyleNeRF and StyleSDF, we adopted the same training settings as used for AFHQv2 training, given that AFHQv2-cats constitutes a subset of AFHQv2.
For VolumeGAN, we followed the same settings as Cats~\cite{Zhang2008CatHD} in pi-gan, including FOV, ray's near/far distances, and camera pose sampling distribution.
For EpiGRAF, we employed the landmark detector\footnote{https://github.com/kairess/cat\_hipsterizer} used in EG3D to label camera poses, while following the guidelines from the EpiGRAF's official repository for other training settings.
The FOV and ray's near/far distances used in EpiGRAF are almost identical to those in pi-gan.

For GIRAFFE-HD on CompCars, we applied transfer-learning from the official checkpoint for $256^2$ resolution to $512^2$ resolution following the authors' guidelines. We trained the model until it achieved the FID reported in the original paper.
\Tref{tab:supp_fid} provides the FIDs we obtained from various sources.

\section{User study}
\label{supp:user}

\begin{figure}[t]
\begin{center}
    \includegraphics[width = 0.9\linewidth]{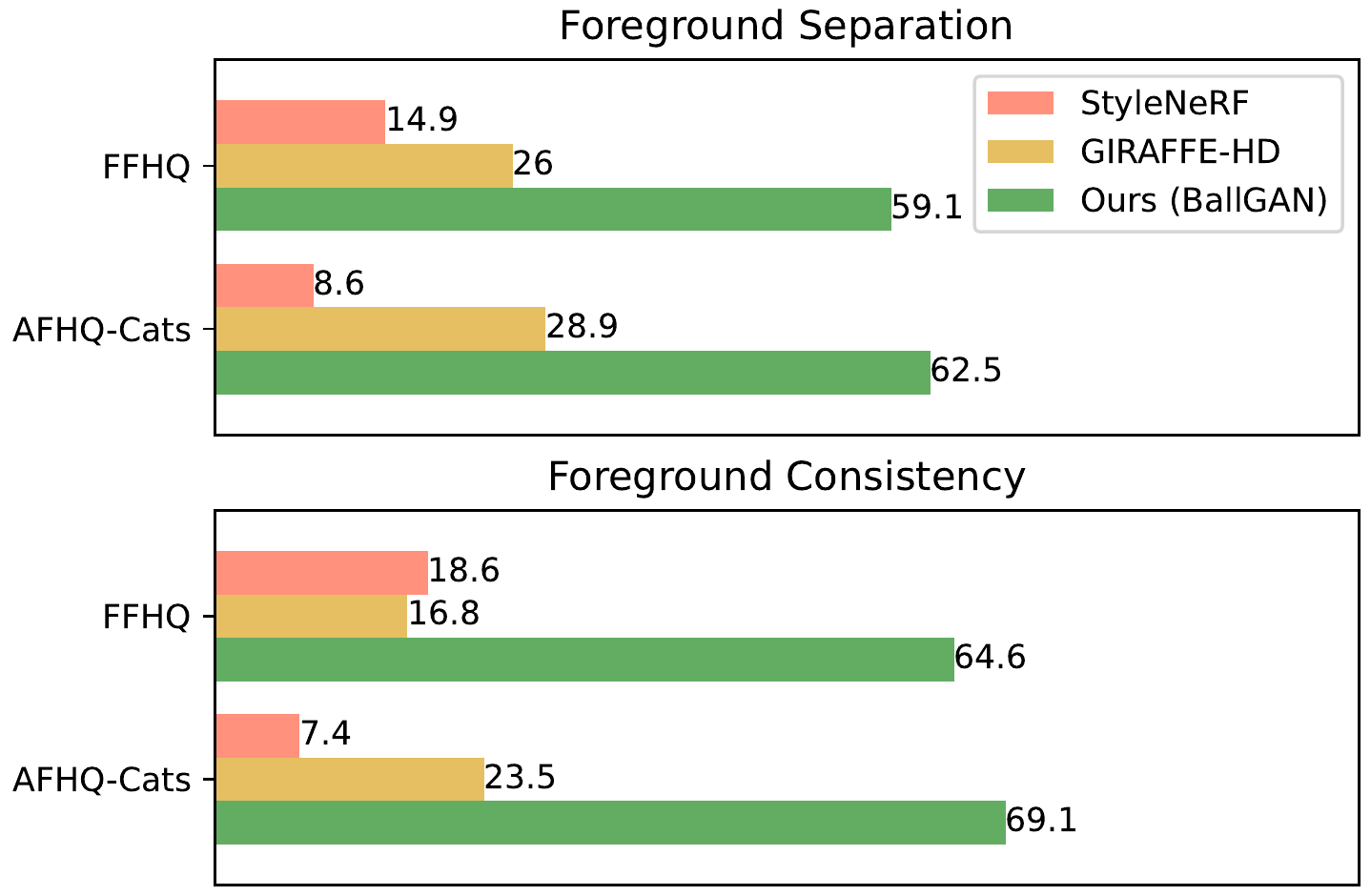}
    \caption{\textbf{User study.}}
    \label{fig:user}
\end{center}
\end{figure}

We asked 57 participants to choose the best model in terms of foreground separation and consistency. We prepared the following questionnaire for our user study in \Fref{fig:user}.
We randomly sampled ten scenes from each method and rendered foregrounds in seven different viewing directions; the entire samples are shown in \Sref{supp:eval}. 
Then we asked 57 participants to answer two questions: (1:Foreground Separation) Which set of foreground fully includes the whole person (or cat) and excludes the background? (2 : Foreground Consistency) Which set of foregrounds is consistent across different views?

\Fref{fig:user} shows that ours outperforms competitors by a large margin with respect to both criteria. See \Sref{supp:eval} for how we prepared images for the user study.

\section{Evaluation protocols}
\label{supp:eval}
We mostly follow the evaluation protocols of EG3D\cite{chan2022efficient}. Below enumerates the protocols.

\myparagraph{Real image inversion}
We use the same configuration of EG3D for pivotal tuning inversion~\cite{roich2022pivotal}.

\myparagraph{ID}
ID measures the cosine similarity of the ArcFace embedding~\cite{deng2019arcface} between different views of the same scene.
For each method, we generate 1000 random scenes in pairs of random poses from the training dataset pose distribution. Then we compute the average.

\myparagraph{Pose}
Pose computes the difference between the intended (input) pose and the synthesized pose, implying how accurately the input poses are reflected in the rendered poses. We sample 1000 latent codes and render them in varying yaws and estimate the resulting yaws with a pre-trained face reconstruction model~\cite{deng2019accurate}. Instead of random yaws, we remove the stochasticity of the evaluation by specifying nine yaw angles evenly separated in [-0.9rad, 0.9rad]. $\pm0.9$rad covers the [0.3, 99.7] percentile of the training dataset's yaw distribution.
We report a mean absolute error (L1) instead of L2 distance to equally capture the error near zero. 

\myparagraph{Depth}
Depth measures the difference between the underlying 3D geometry (volume-rendered depth) and the rendered image. We consider depth maps of rendered images in frontal views of 1000 samples estimated by a pre-trained 3D face reconstruction model~\cite{deng2019accurate} as pseudo ground truth. The depth maps are normalized to compute their mean squared error. 

\begin{figure}
\begin{center}
    \centering
    \includegraphics[width=\linewidth]{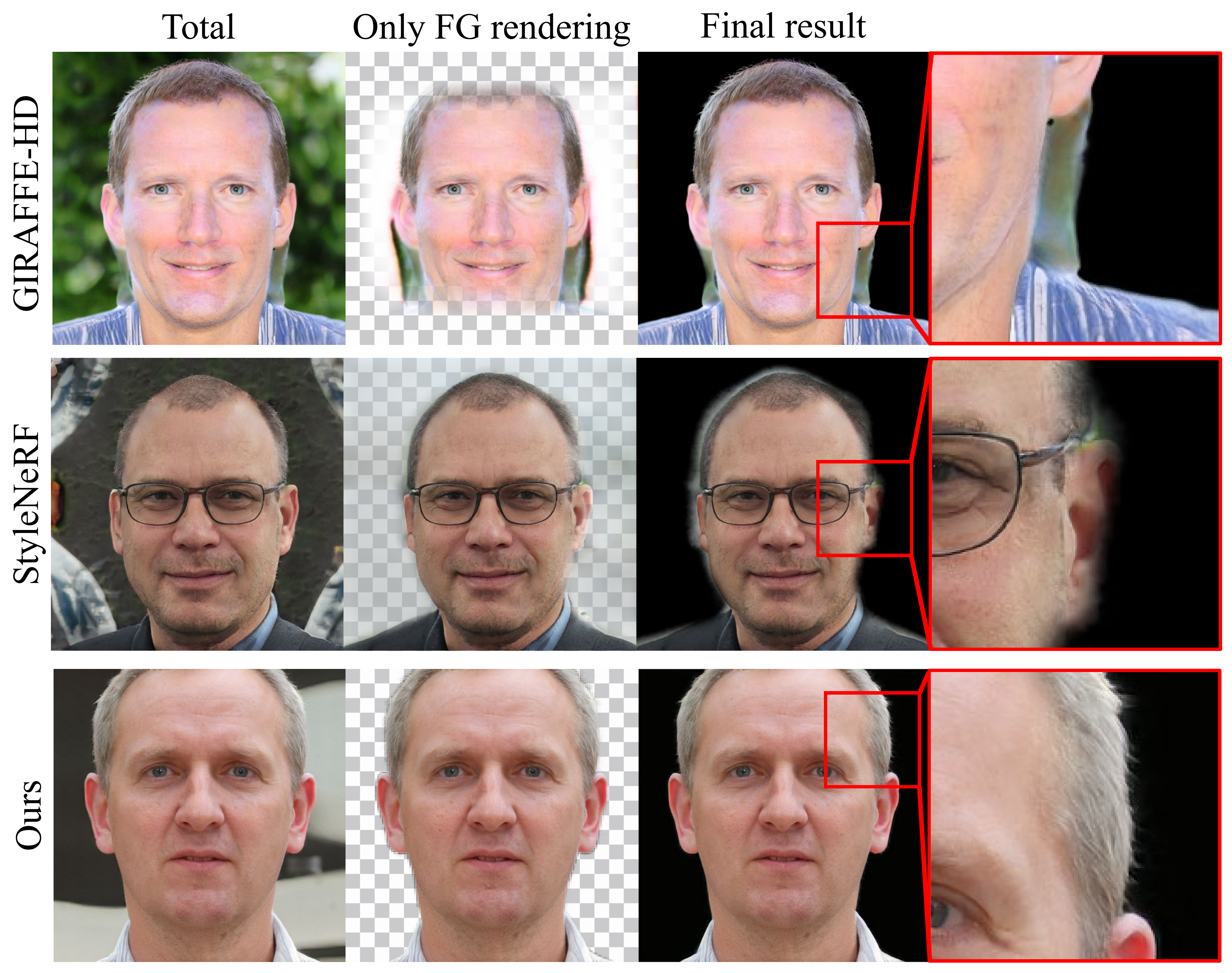}
    \caption{\textbf{Foreground separation examples.} The densities along a ray do not sum to one in GIRAFFE-HD and StyleNeRF. Hence, we apply postprocessing to compare their full potential for separation. Ours does not require such postprocessing. The rightmost column shows zoomed-in images of \textcolor{red}{red} box regions for detailed comparison.} 
    \label{fig:fg_postprocessing}
\end{center}
\end{figure}

\myparagraph{Foreground separation}
We describe the procedure to obtain the foreground image used in \Sref{exp:separation}. 
Although our goal is to compare the separation of foreground and background in the 3D space, it is prohibitive to visualize the separation in 3D space on paper or screen. Therefore, we visualize by separately synthesizing the foreground scene for each method. 
Note that GIRAFFE-HD produces extra alpha masks in 2D space. We visualize their foreground part with their alpha masks to demonstrate their best performance. Their foreground densities are only in the central region of the image canvas, and their aggregated densities do not match the shape of the salient object. 
For StyleNeRF, the foreground densities along the ray do not sum to one, \ie, the foreground is semi-transparent. 
Therefore, we manually searched for a density threshold that best divides the foreground region for each image.
Ours do not require such workarounds as the foreground densities aggregate to one along the rays well on the foreground regions. \Fref{fig:fg_postprocessing} provides examples.



\section{Detailed qualitative comparison}
\label{supp:qual compare}

\begin{figure}
\begin{center}
    \centering
    \includegraphics[width=\linewidth]{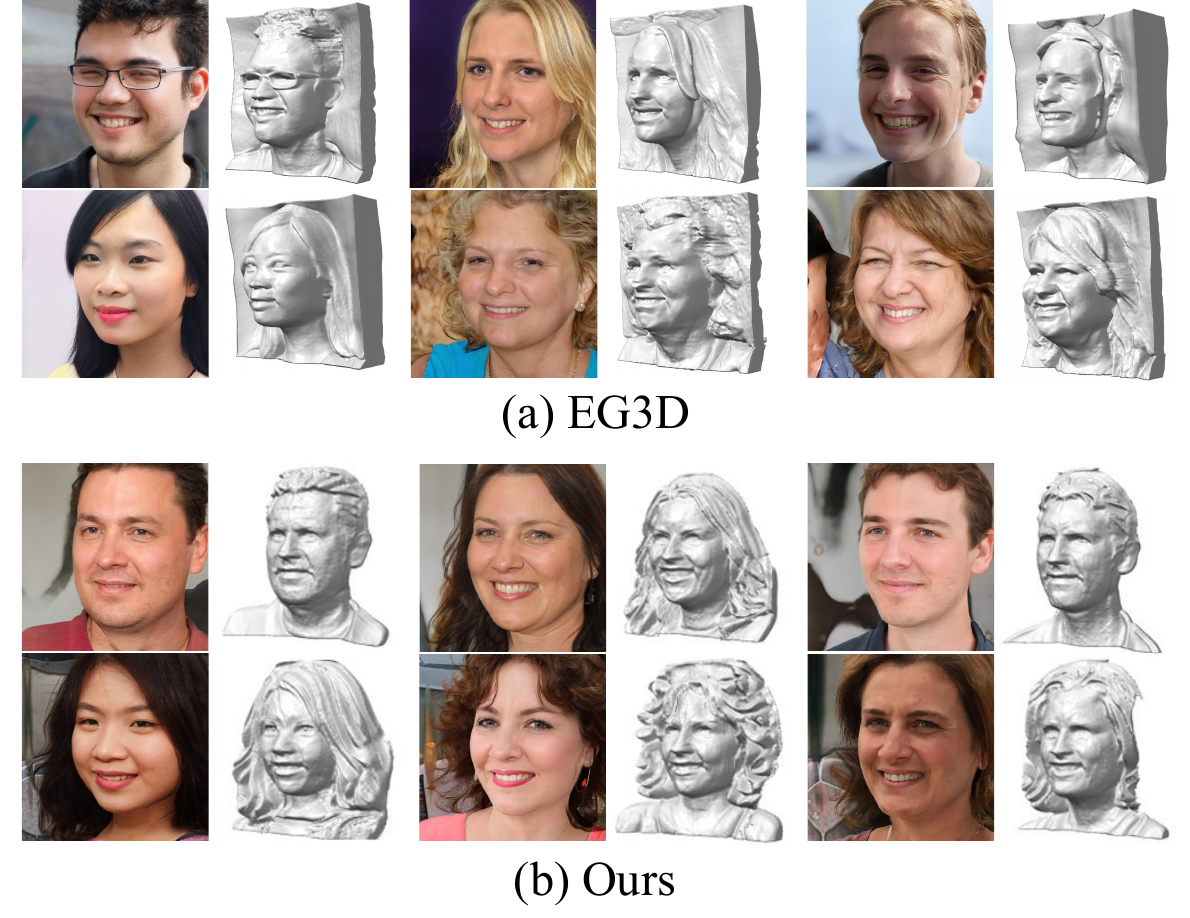}
    \caption{\textbf{3D geometry comparison between EG3D and BallGAN}}
    \label{fig:vsEG3D}
\end{center}
\end{figure}

We only visualize the foreground meshes in \Fref{fig:BallGAN-S}, \Fref{fig:geometry}, \Fref{fig:vsEG3D}, and \Fref{fig:styleclip} for methods that separately model on foreground and background.
\Fref{fig:teaser}, \Fref{fig:degenerate} and \Fref{fig:varyingview} show the full 3D scene, including both foreground and background.
As EG3D does not separate foreground and background, the full 3D geometry is visualized on all mesh figures.

However, we only visualize the foreground mesh of StyleNeRF in \Fref{fig:varyingview} as we discover that the background densities of StyleNeRF are close to zero, thus negligible. Yet, the background appears on rendered images of StyleNeRF as the last sample on the background ray is set to have an alpha value of 1 before volume rendering, i.e., the alpha value for the last sample is tweaked to 1 regardless of the actual density produced by the background NeRF. 

Despite the sole visualization of foreground mesh for StyleNeRF in \Fref{fig:varyingview}, densities accountable for background is noticeable on StyleNeRF's mesh for AFHQv2-Cats. This shows the case of the background being erroneously modeled through the foreground.

\begin{figure}[t]
\begin{center}
    \centering
    \includegraphics[width=\linewidth]{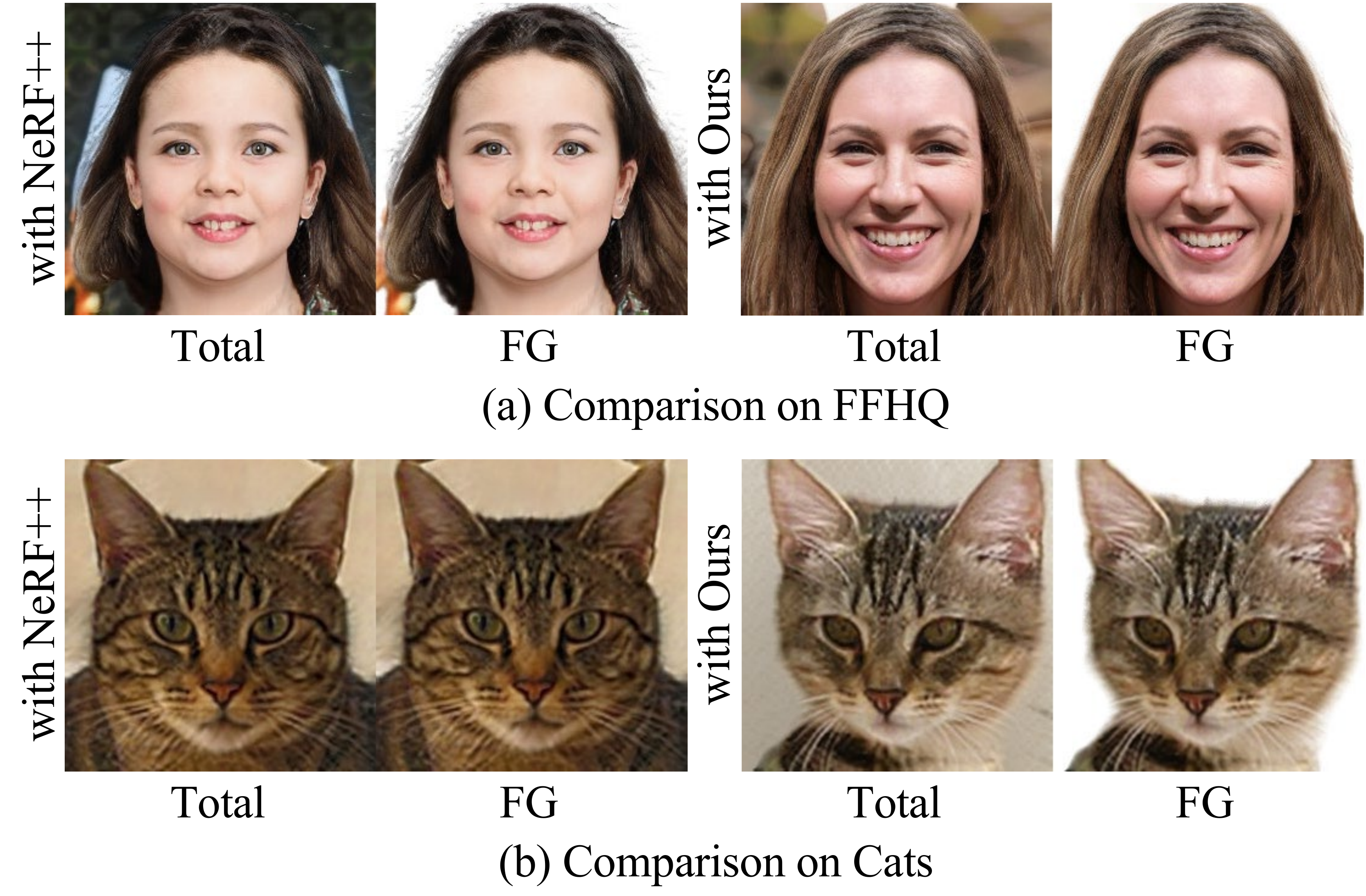}
    \caption{\textbf{Comparison of foreground and background separation with EpiGRAF backbone} NeRF++ BG struggles on hair, shoulder, and cat. Our BG excels in all cases.}
    \label{fig:EpiGRAF_separation}
\end{center}
\end{figure}

EpiGRAF employs NeRF++'s inverse sphere parameterization for the background, the same as StyleNeRF.
\Fref{fig:EpiGRAF_separation} shows a comparison between our background representation and NeRF++ when using EpiGRAF as the backbone. The term "with NeRF++" refers to the original EpiGRAF, while "with Ours" indicates the model where our sphere background representation is applied to EpiGRAF's foreground representation. 
Except for the background representation, all settings remain  the same and adhere to the guidelines provided in the official repository.

In FFHQ, EpiGRAF with Ours separates the FG cleaner. 
On the Cats~\cite{Zhang2008CatHD} dataset, which contains a significant amount of fine-grained details, EpiGRAF with NeRF++ fails to separate the FG and BG, whereas EpiGRAF with Ours shows clear separation.

\begin{figure}
\begin{center}
    \centering
    \includegraphics[width=0.95\linewidth]{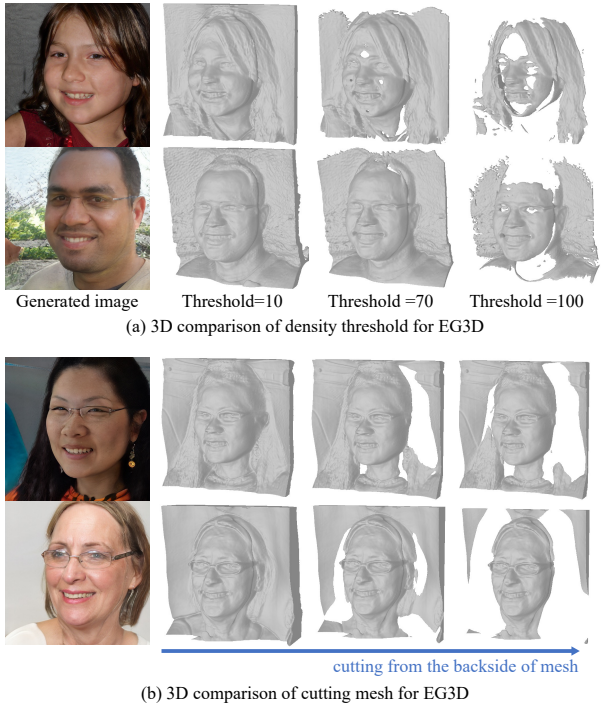}
    \caption{\textbf{Difficulty of separating foreground in EG3D} (a) The background cannot be removed by thresholding density, i.e., the foreground is cut off before the background is fully removed. (b) As the background wall has a concave shape and is not always behind the foreground, clipping with depth tends to carve out the foreground before full background removal.}
    \label{fig:EG3D_fgseperation}
\end{center}
\end{figure}

\section{More comparison with EG3D}
\label{supp:eg3d_comparison}

EG3D does not separately model foreground and background. 
\Fref{fig:vsEG3D} highlights the drawback of this representation for learning 3D scenes.
The ears and hair in 3D space are attached to the background. Some parts of the hair are flat and lack curls. 
In contrast, ours separates the hair from the background and correctly models the 3D geometry of the hair that matches the 2D observation.

\Fref{fig:EG3D_fgseperation} shows that foreground separation is not straightforward in EG3D's 3D space. 
Thresholding the density or carving the mesh from the back does not correctly separate the foreground, and damages the facial/hair regions first. 
This demonstrates that the foreground and background must be perfectly separated at the representation level.

\section{Detailed multi-view comparison}
\label{supp:comparison}
\begin{figure*}[ht]
  \centering
  \begin{subfigure}{\linewidth}
    \includegraphics[clip=true,width=1.0\linewidth]{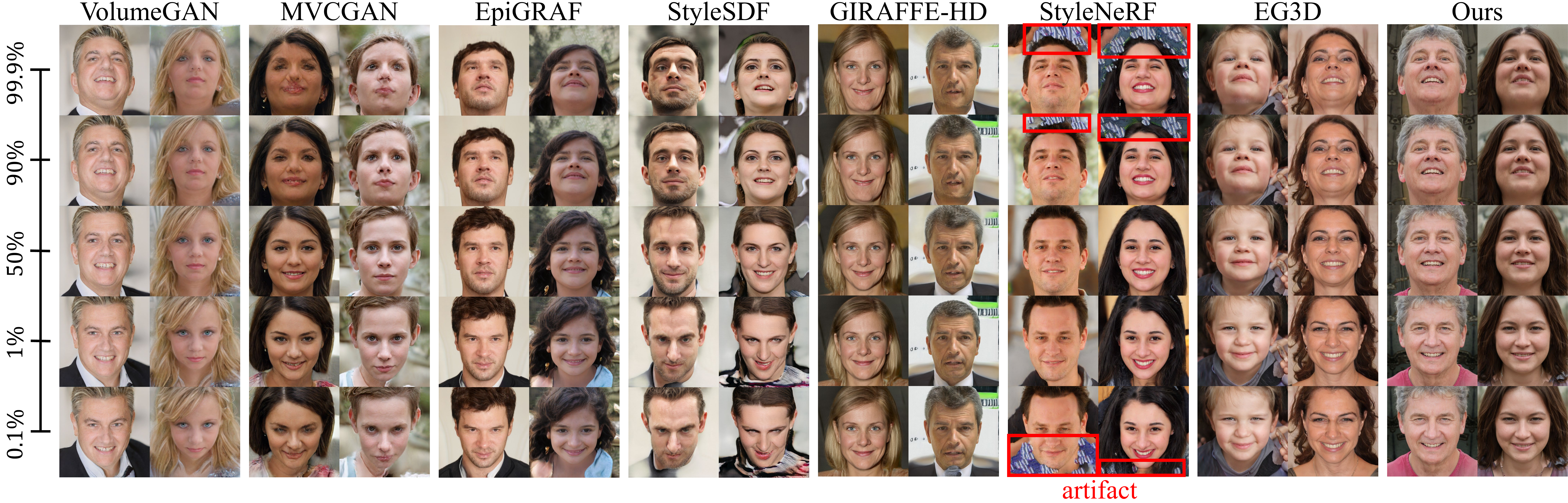}
    \caption{Multi-view comparison with varying pitches}
    \label{fig:pitch}
  \end{subfigure}
  \vspace{5mm}
  \hfill
  \begin{subfigure}{\linewidth}
    \includegraphics[clip=true,width=1.0\linewidth]{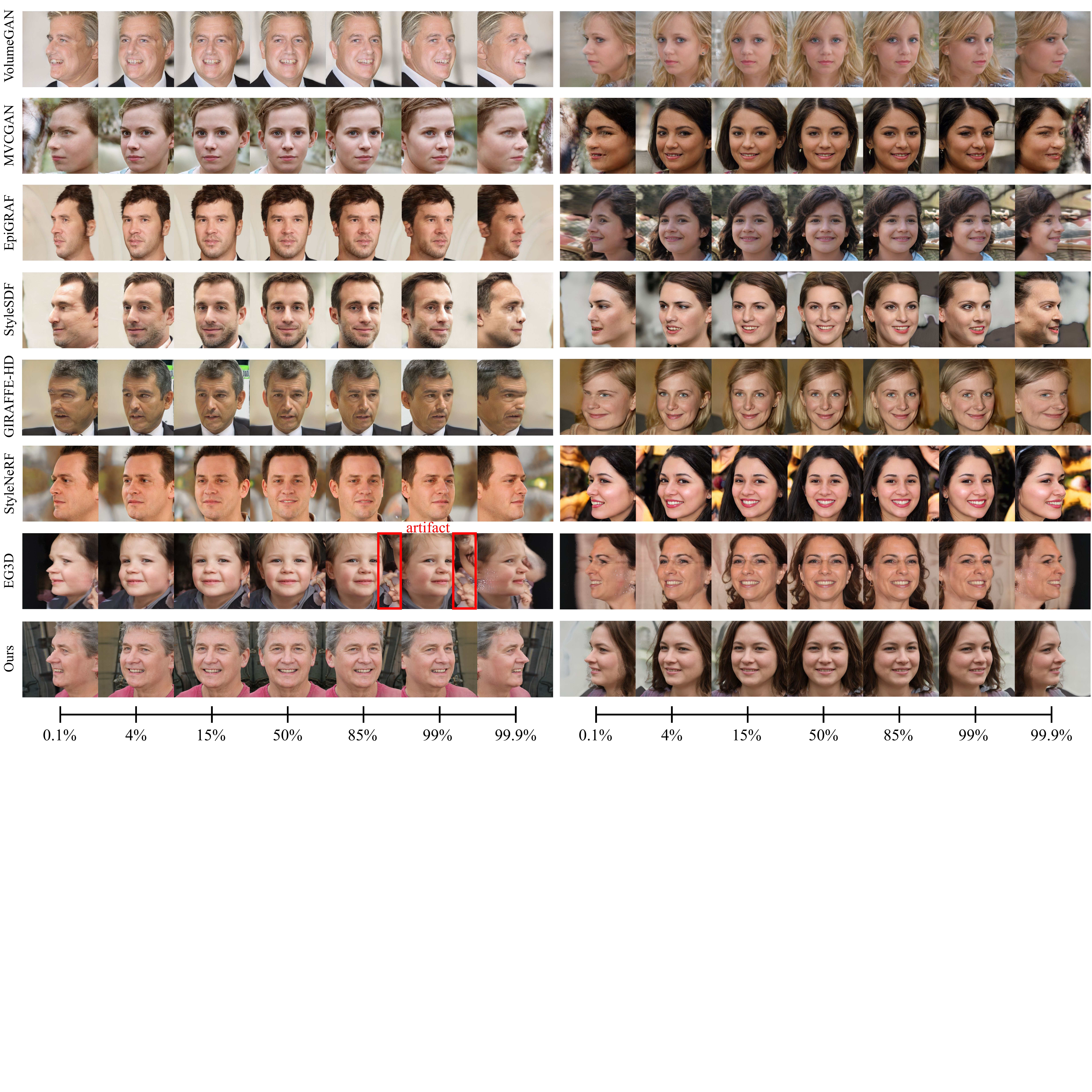}
    \caption{Multi-view comparison with varying yaws}
    \label{fig:yaw}
  \end{subfigure}
  \caption{\textbf{Multi-view comparison in various poses on FFHQ.} Percentile for camera pitch and yaw in training distribution are shown on the left side of \subref{fig:pitch} and below for \subref{fig:yaw}.}
  \label{fig:multi-view}
\end{figure*}

\Fref{fig:pitch} and \Fref{fig:yaw} provide qualitative comparisons with varying camera poses. 
As FFHQ dataset mainly consists of frontal views, the competitors produce artifacts or show multi-view inconsistency. On the other hand, BallGAN produces images that are multi-view consistent and free from artifacts even in extreme camera poses.
\section{Uncurated samples}
\label{supp:uncuratedimages}
\begin{figure*}[ht]
  \centering
  \begin{subfigure}{\linewidth}
    \includegraphics[clip=true,width=1.0\linewidth]{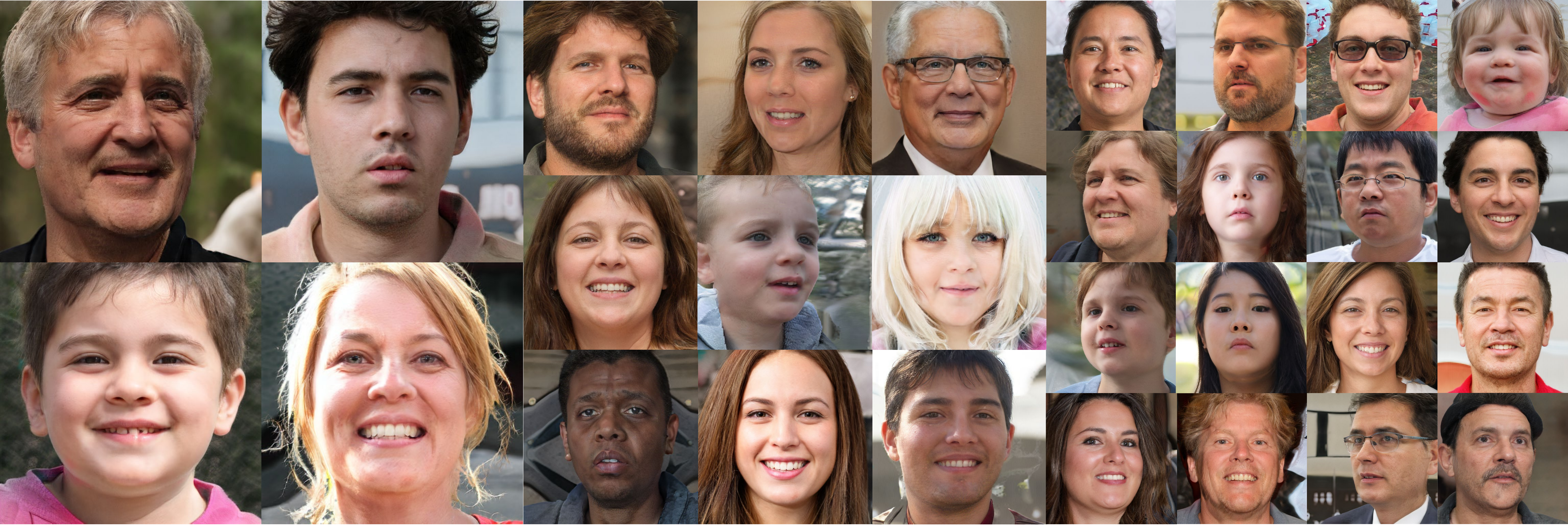}
    \caption{Uncurated samples of FFHQ.}
    \label{fig:uncurate_FFHQ}
  \end{subfigure}
  \begin{subfigure}{\linewidth}
    \includegraphics[clip=true,width=1.0\linewidth]{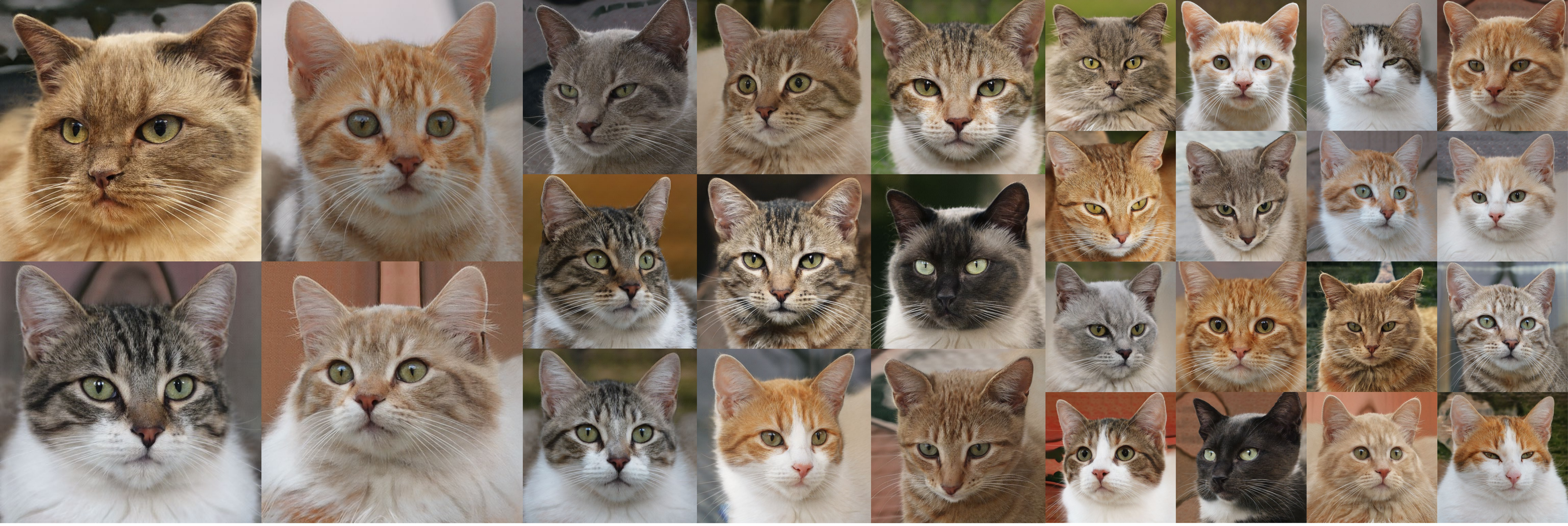}
    \caption{Uncurated samples of AFHQv2-Cats.}
    \label{fig:uncurate_AFHQ}
  \end{subfigure}
   \begin{subfigure}{\linewidth}
    \includegraphics[clip=true,width=1.0\linewidth]{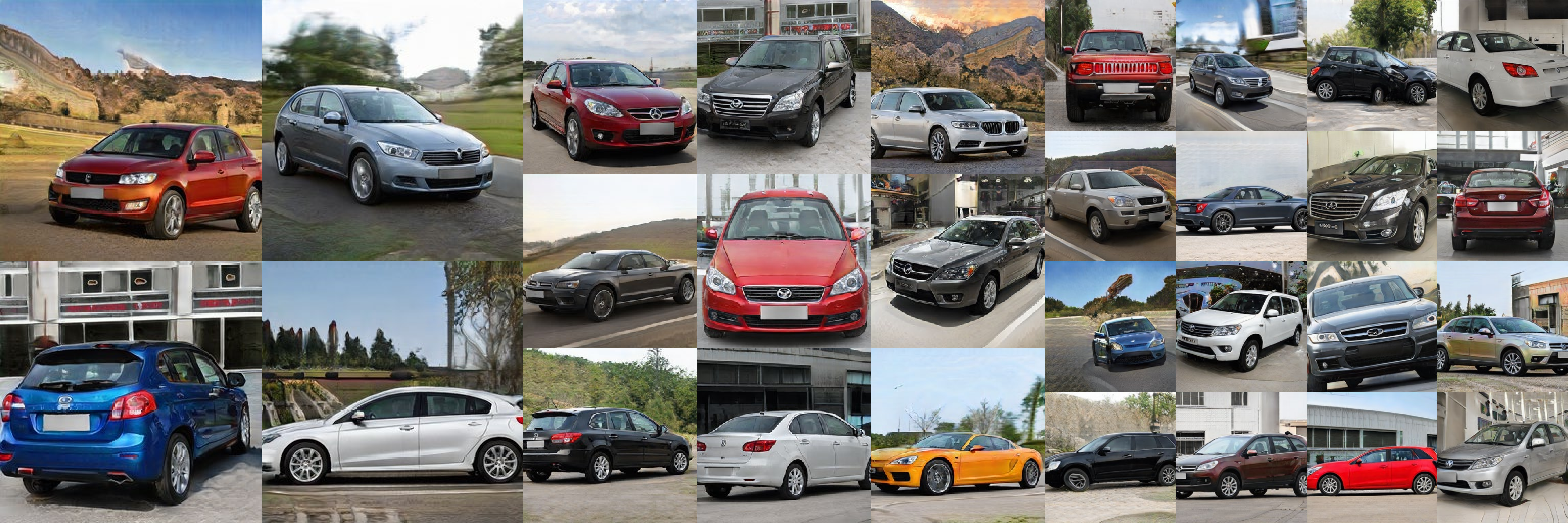}
    \caption{Uncurated samples of CompCars.} 
    \label{fig:uncurate_CompCars}
  \end{subfigure}
  \caption{\textbf{Uncurated samples on the FFHQ, AFHQv2-Cats, and CompCars.} Camera poses are randomly chosen from each training distribution. \subref{fig:uncurate_FFHQ} and \subref{fig:uncurate_AFHQ} show outputs of BallGAN. \subref{fig:uncurate_CompCars} is outputs from BallGAN-S.}
  \label{fig:uncurated}
\end{figure*}

\Fref{fig:uncurated} provides uncurated samples of our method.


\clearpage
\clearpage

\end{document}

%% file: tables/3D.tex
\begin{table}[t]
\centering
\resizebox{\columnwidth}{!}{
\begin{tabular}{lcllllllll}
\toprule
           & \multicolumn{9}{c}{FFHQ $512^2$}                                                                                                                                                                                           \\
           & \multicolumn{3}{c}{ID $\uparrow$}                                      & \multicolumn{3}{c}{Pose $\downarrow$}                                   & \multicolumn{3}{c}{Depth $\downarrow$}                                  \\ \midrule
MVCGAN     &                      & \multicolumn{1}{c}{0.58} & \multicolumn{1}{c}{} & \multicolumn{1}{c}{} & \multicolumn{1}{c}{0.014} & \multicolumn{1}{c}{} & \multicolumn{1}{c}{} & \multicolumn{1}{c}{0.123} & \multicolumn{1}{c}{} \\
VolumeGAN  &                      & \multicolumn{1}{c}{0.63} & \multicolumn{1}{c}{} & \multicolumn{1}{c}{} & \multicolumn{1}{c}{0.025} & \multicolumn{1}{c}{} & \multicolumn{1}{c}{} & \multicolumn{1}{c}{0.020} & \multicolumn{1}{c}{} \\
StyleSDF   &                      & \multicolumn{1}{c}{0.50} & \multicolumn{1}{c}{} & \multicolumn{1}{c}{} & \multicolumn{1}{c}{0.010} & \multicolumn{1}{c}{} & \multicolumn{1}{c}{} & \multicolumn{1}{c}{0.016} & \multicolumn{1}{c}{} \\
EpiGRAF   &                      & \multicolumn{1}{c}{0.71} & \multicolumn{1}{c}{} & \multicolumn{1}{c}{} & \multicolumn{1}{c}{0.013} & \multicolumn{1}{c}{} & \multicolumn{1}{c}{} & \multicolumn{1}{c}{0.143} & \multicolumn{1}{c}{} \\
EG3D       & \multicolumn{1}{l}{} & 0.71                     &                      &                      & 0.007                     &                      &                      & 0.011                     &                      \\ \midrule
GIRAFFE-HD & \multicolumn{1}{l}{} & 0.69                     &                      &                      & 0.064                     &                      &                      & 0.058                     &                      \\
StyleNeRF  & \multicolumn{1}{l}{} & 0.64                     &                      &                      & 0.018                     &                      &                      & 0.013                     &                      \\
Ours       & \multicolumn{1}{l}{} & \textbf{0.75}                     &                      &                      & \textbf{0.005}                     &                      &                      & \textbf{0.008}                     &                      \\ \bottomrule
\end{tabular}
}
\vspace{-2mm}
\captionof{table}{Quantitative evaluation on 3D geometry. We report identity consistency (ID), pose accuracy, and depth errors for FFHQ. Our method outperforms baselines in all metrics of 3D-awareness.}
\label{tab:3D}
\end{table}

%% file: tables/colmap.tex
\begin{table}[t]
\newcommand{\himg}{0.242}
\centering
\small
\begin{minipage}{\linewidth}
    \centering
    \makebox[\himg\linewidth][c]{{GIRAFFE-HD}}
    \makebox[\himg\linewidth][c]{{StyleNeRF}}
    \makebox[\himg\linewidth][c]{{EG3D}}
    \makebox[\himg\linewidth][c]{{Ours}~}   
    \includegraphics[width=\himg\linewidth]{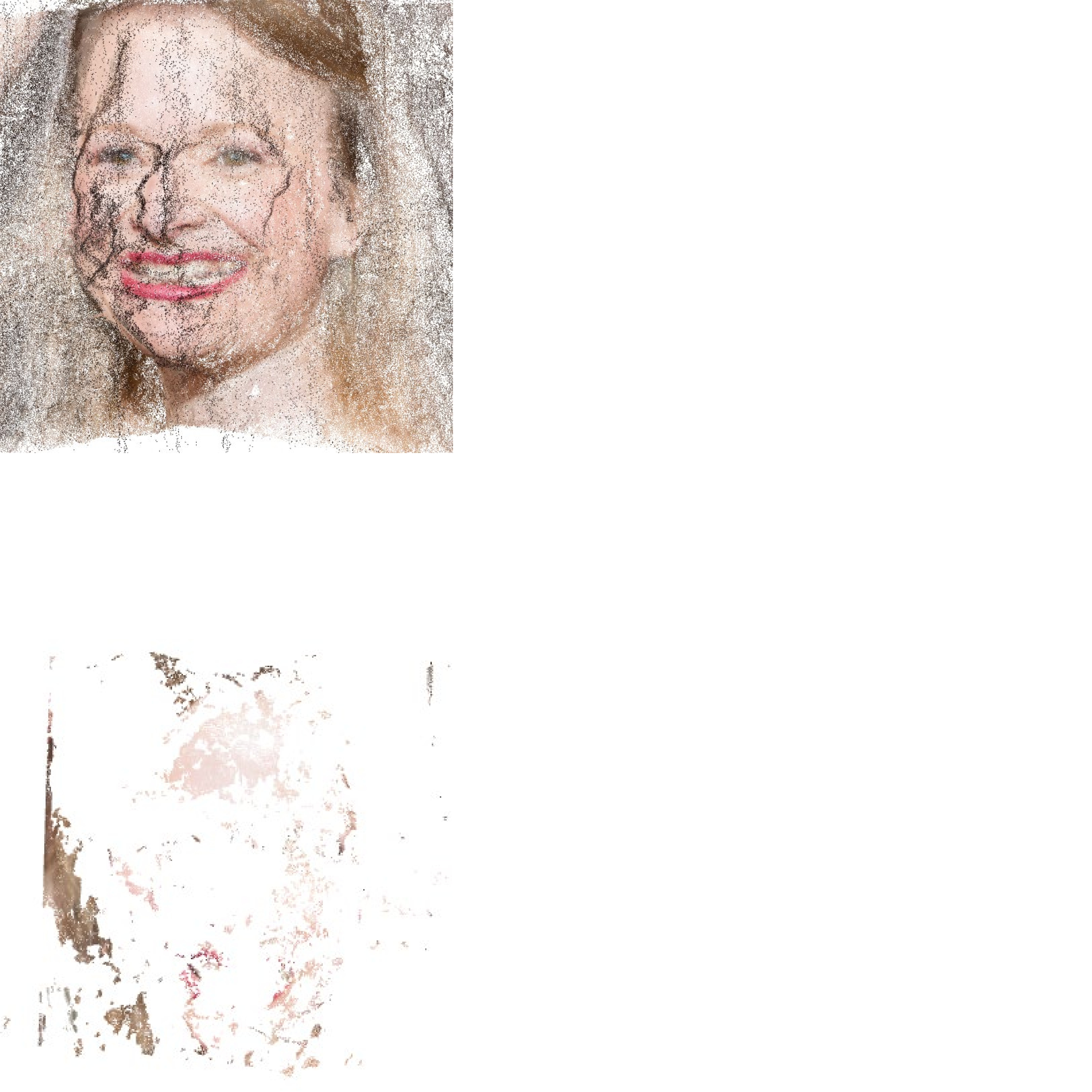}
    \includegraphics[width=\himg\linewidth]{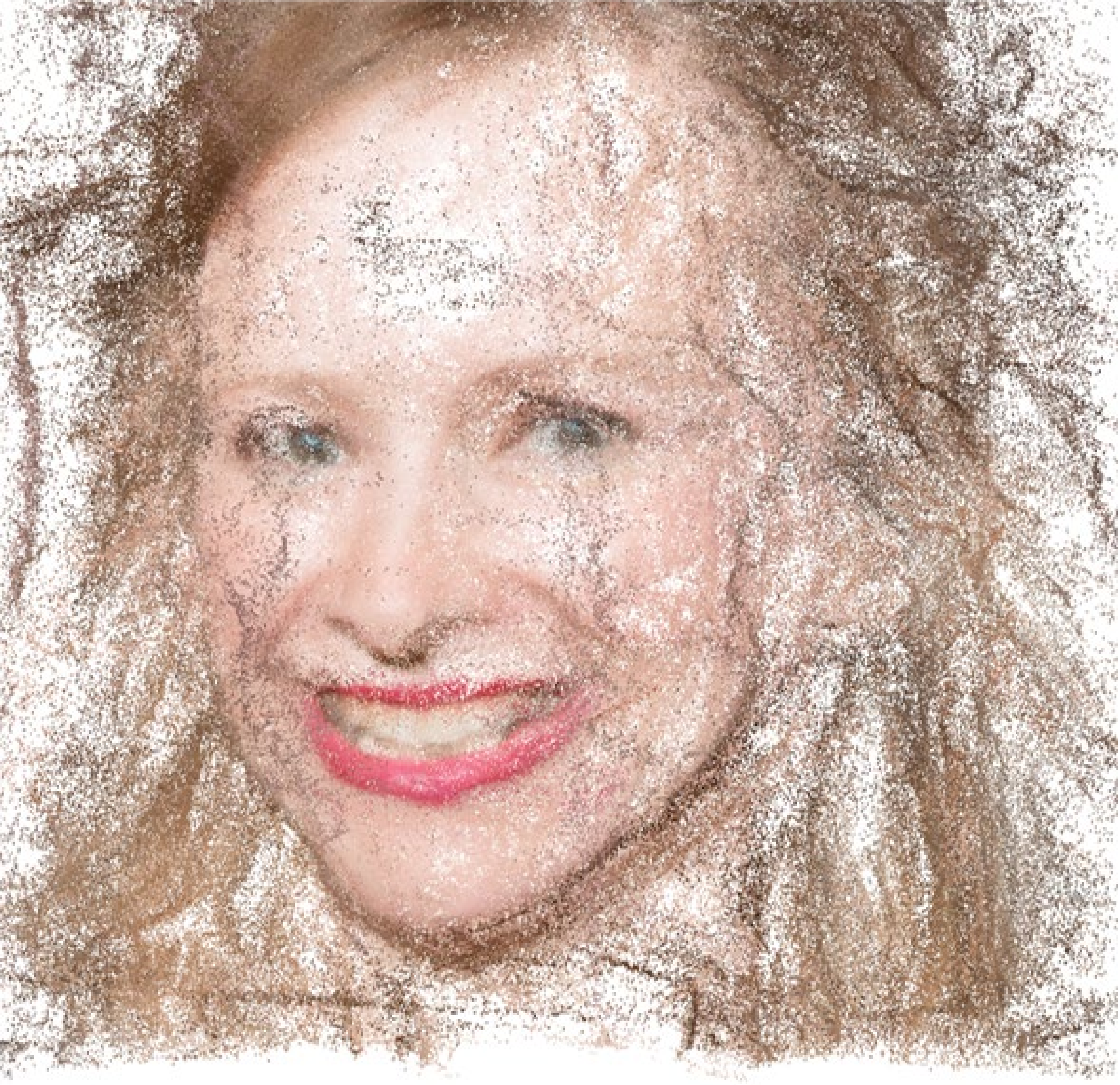}
    \includegraphics[width=\himg\linewidth]{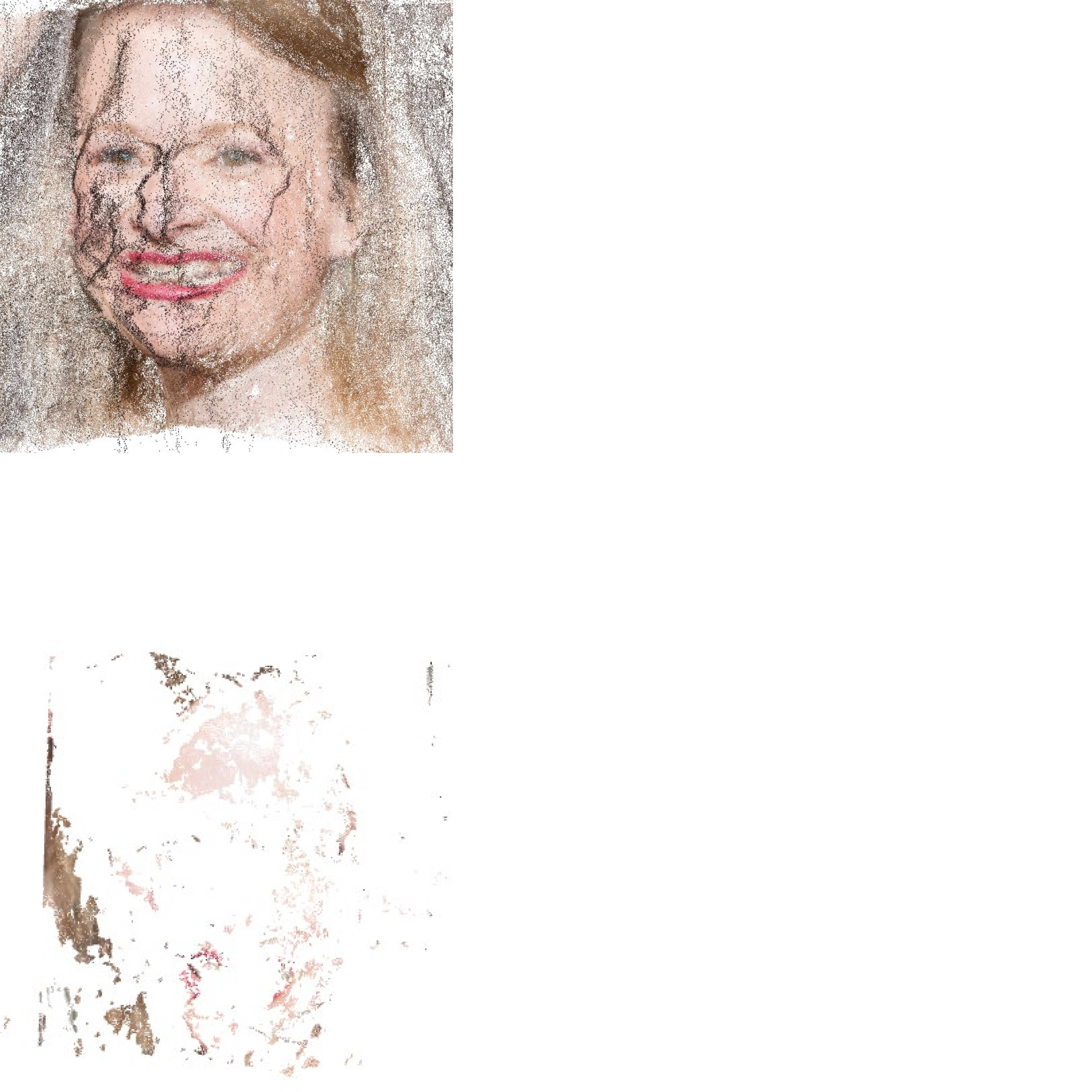}
    \includegraphics[width=\himg\linewidth]{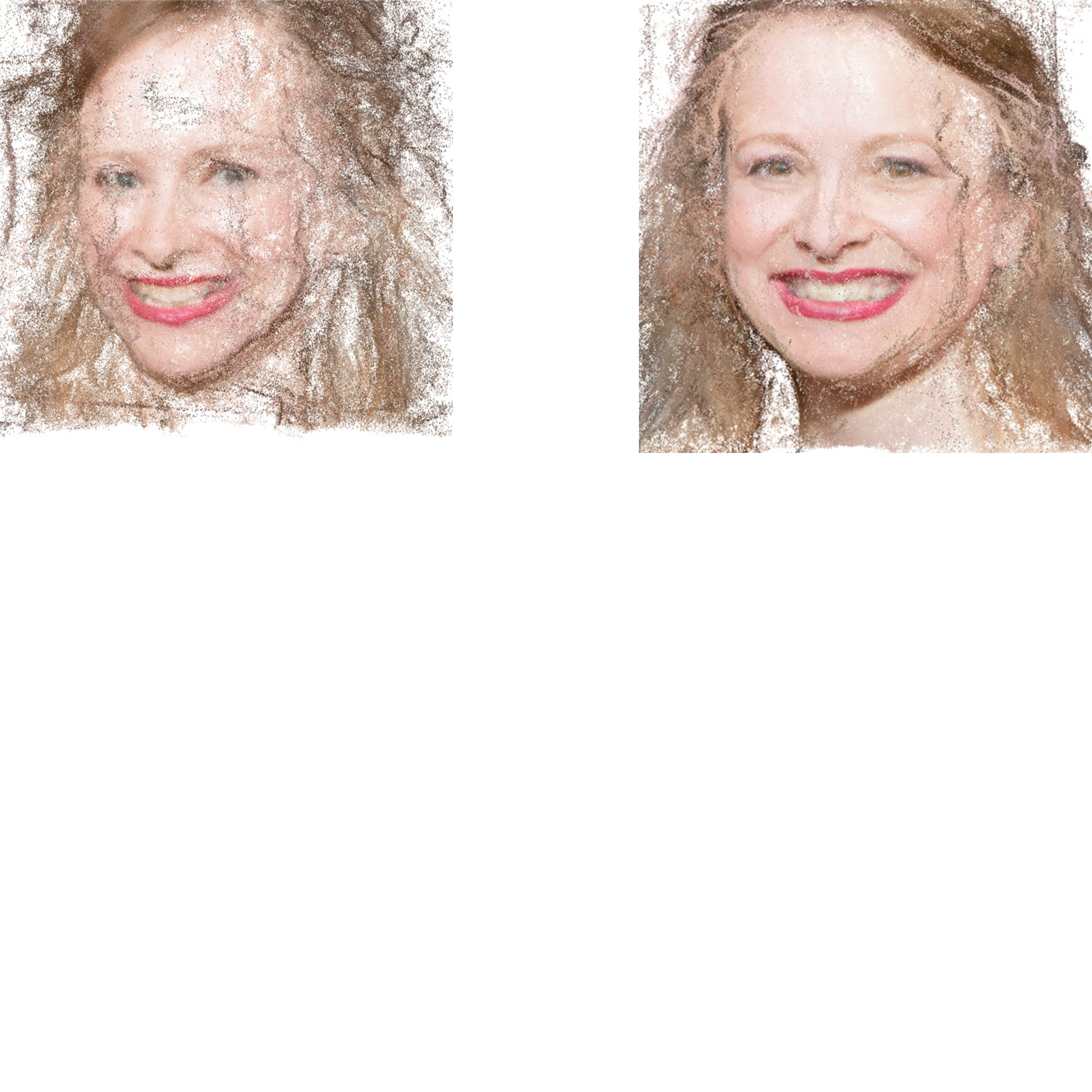}
\end{minipage}
\small

\resizebox{\columnwidth}{!}{
\begin{tabular}{lcccc}
\toprule
\textbf{Method} & \footnotesize{GIRAFFE-HD} & \footnotesize{StyleNeRF} & EG3D & Ours \\
\cmidrule(lr){1-1}
\cmidrule(lr){2-5}

$\#$ of rec. ($10^4$) \ \ 
    & $17 \pm 2.3$  & $53 \pm 8.4$  
    & $78 \pm 5.5$   & $79 \pm 5.0$
\\[0.3mm]

\bottomrule
\end{tabular}
}
\vspace{-2mm}
\caption{\textbf{COLMAP point cloud reconstruction} is performed using 128 views in $[-\pi/2, \pi/2]$ from the generated scene for each model. A higher number of reconstructed points indicates better multi-view consistency.}
\label{tab:colmap}
\vspace{-3mm}
\end{table}

%% file: tables/imagequality.tex


\begin{table}[t]
\centering
\resizebox{\columnwidth}{!}{
\small
\begin{tabular}{@{\hspace{1mm}}llrrr@{\hspace{1mm}}} 
\toprule
 \begin{tabular}[c]{@{}l@{}}Sep. \\ FG/BG\end{tabular} & \multicolumn{1}{c}{} & \multicolumn{1}{c}{\begin{tabular}[c]{@{}c@{}}FFHQ \\ \textbf{$512^2$}\end{tabular}} & \multicolumn{1}{c}{\begin{tabular}[c]{@{}c@{}} AFHQv2-Cats  \\ \textbf{$512^2$}\end{tabular}} & \multicolumn{1}{c}{\begin{tabular}[c]{@{}c@{}}CompCars \\ \textbf{$256^2$}\end{tabular}} \\ 
\midrule
\multirow{4}{*}{\xmark} & MVCGAN & 13.4$^\dagger$ & 26.57$^\ddagger$ & - \\
 & VolumeGAN & 15.74 & 44.55 & 12.9$^\dagger$ \\
 & StyleSDF & 19.56 & 19.44 & - \\
 & EpiGRAF & 9.92$^\dagger$  & 6.46 & - \\
 & EG3D & \textbf{4.7}$^\dagger$ & \textbf{2.77}$^\dagger$ & N/A \\ 
\midrule

\multirow{3}{*}{\cmark} & GIRAFFE-HD & 6.47 & 7.33 & \uline{7.1}$^\ddagger$ \\
 & StyleNeRF & 10.51$^\ddagger$ & 21.56 & 8$^\dagger$ (284$\pm$96) \\
 & Ours & \uline{5.67} & \uline{4.72} & \textbf{4.26} \\
\bottomrule
\end{tabular}
}
\vspace{-2.5mm}
\captionof{table}{Quantatitive comparison using FID~\cite{heusel2017gans} on three datasets. $\dagger$ denotes the reported FID, and $\ddagger$ denotes the FID calculated by the official checkpoint. In other cases, we train each baseline using their official codes. In the case of StyleNeRF on CompCars, we report FID of diverged models over 3 experiments in the parenthesis. 
N/A denotes the model can not be trained.
Bold and underline indicate the best and second-best performance. 
Our method shows the best score in CompCars and comparable scores with EG3D. } 
\vspace{-3mm}
\label{tab:quals}
\end{table}

%% file: tables/supp_abl.tex
\begin{table}[t]
\centering
\begin{tabular}{lccrr} 
\hline
& \multicolumn{2}{c}{configuration}   &  \\
& $\mathcal{L}_\text{fg}$ & $\mathcal{L}_\text{bg}$   & FID \\
\hline
\multirow{4}{*}{stage 1} & - & - & 7.87 \\
& \ding{51} & - & 6.82 \\
& - & \ding{51} & 7.88 \\
& \ding{51} & \ding{51} & 6.13 \\
\hline
\end{tabular}
\caption{\textbf{Ablation study on regularization.} This ablation study is conducted with batch size 16 due to the resource shortage. FIDs do not match the main results.} 
\label{tab:qual_abl}
\end{table}

%% file: tables/supp_bg_arc.tex
\begin{table}[h]
\centering
\begin{tabular}{ccc}
\hline
\multicolumn{1}{l}{} & \multicolumn{1}{l}{input channel} & \multicolumn{1}{l}{output channel} \\ \hline
\textit{PE}          & 2                                 & 40                                 \\
$\vg^{1}_{\wbg}$     & 40                                & 64                                 \\
$\vg^{2}_{\wbg}$     & 64                                & 64                                 \\
$\vg^{3}_{\wbg}$     & 64                                & 64                                 \\
$\vg^{4}_{\wbg}$     & 64                                & 64                                 \\
$\vg^{5}_{\wbg}$     & 64                                & 32                                 \\ \hline
\end{tabular}
\caption{\textbf{Detail of background network.} \textit{PE} means positional encoding $\zeta$, not a layer.} 
\label{tab:bg_arc}
\end{table}

%% file: tables/supp_fid.tex
\begin{table}[h]
\centering
\resizebox{1\columnwidth}{!}
{
\begin{tabular}{lrrrr}
\cline{1-5}\cline{1-5}
\multicolumn{1}{c}{} & \multicolumn{3}{c}{FFHQ $512^2$} & \multicolumn{1}{c}{FFHQ other res.}\\
              & \multicolumn{1}{l}{reported} & \multicolumn{1}{c}{reproduced} & \multicolumn{1}{c}{official ckpt.} & \multicolumn{1}{c}{reported} \\ \hline
GRAM               & -    & -    &  -   & ($256^2$) 29.8\\
MVCGAN               & \textbf{13.4}     & -    & 21.3   \\
VolumeGAN            & -    & \textbf{15.7}    & -   & ($256^2$) 9.1\\
StyleSDF             & -    & \textbf{19.5}    & -   & ($256^2$) 11.5\\
EpiGRAF              & \textbf{9.9}    &  -    & -   & ($256^2$) 9.7\\
EG3D                 & \textbf{4.7}    & 4.7    & -  & \\ \hline
GIRAFFE-HD           & -    & \textbf{6.4}    & -   & ($1024^2$) 10.13\\
StyleNeRF            & 13.2    & -    & \textbf{10.5} &  \\
Ours            & \textbf{5.64}  \\
\cline{1-5}\cline{1-5}
\end{tabular}
}
\caption{\textbf{FIDs of competitors from various sources.} We report the best FID among the reported, reproduced and official checkpoint for each model with $512^2$ resolutions in \Tref{tab:quals}.}
\label{tab:supp_fid}
\end{table}